\documentclass[10pt,journal,compsoc]{IEEEtran}
\usepackage[linesnumbered,ruled]{algorithm2e}
\usepackage{amsmath,amsfonts}
\usepackage{algorithmic}
\usepackage{array}
\usepackage[caption=false,font=normalsize,labelfont=sf,textfont=sf]{subfig}
\usepackage{textcomp}
\usepackage{stfloats}
\usepackage{url}
\usepackage{verbatim}
\usepackage{graphicx}
\usepackage{float}
\usepackage{enumitem}
\usepackage{hyperref}
\hypersetup{colorlinks=true, urlcolor=blue}
\usepackage{cite}
\usepackage{cuted}
\usepackage{booktabs}
\usepackage{multirow}
\usepackage{makecell}
\usepackage{diagbox}
\def\BibTeX{{\rm B\kern-.05em{\sc i\kern-.025em b}\kern-.08em
    T\kern-.1667em\lower.7ex\hbox{E}\kern-.125emX}}
\usepackage{balance}
\begin{document}
\title{Robust Phase-Shifting Profilometry for Arbitrary Motion}

\author{Geyou Zhang, Kai Liu,~\IEEEmembership{Senior Member,~IEEE}, Ao Li, and Ce Zhu*,~\IEEEmembership{Fellow,~IEEE} 
\thanks{G. Zhang, A. Li, and C. Zhu are with the School of Information and Communication Engineering, University of Electronic Science and Technology of China, Chengdu, China, 611731.}
\thanks{K. Liu is with College of Electrical Engineering, Sichuan University, Chengdu, China, 610065.}
\thanks{Corresponding author: C. Zhu, Email: eczhu@uestc.edu.cn.}}


\IEEEtitleabstractindextext{
\begin{abstract}
Phase-shifting profilometry (PSP) enables high-accuracy 3D reconstruction but remains highly susceptible to object motion. Although numerous studies have explored compensation for motion-induced errors, residual inaccuracies still persist, particularly in complex motion scenarios. In this paper, we propose a robust phase-shifting profilometry for arbitrary motion (RPSP-AM), including six-degrees-of-freedom (6-DoF) motion (translation and rotation in any direction), non-rigid deformations, and multi-target movements, achieving high-fidelity motion-error-free 3D reconstruction. We categorize motion errors into two components: 1) ghosting artifacts induced by image misalignment, and 2) ripple-like distortions induced by phase deviation. To eliminate the ghosting artifacts, we perform pixel-wise image alignment based on dense optical flow tracking. To correct ripple-like distortions, we propose a high-accuracy, low-complexity image-sequential binomial self-compensation (I-BSC) method, which performs a summation of the homogeneous fringe images weighted by binomial coefficients, exponentially reducing the ripple-like distortions with a competitive computational speed compared with the traditional four-step phase-shifting method. Extensive experimental results demonstrate that, under challenging conditions such as 6-DoF motion, non-rigid deformations, and multi-target movements, the proposed RPSP-AM outperforms state-of-the-art (SoTA) methods in compensating for both ghosting artifacts and ripple-like distortions. Our approach extends the applicability of PSP to arbitrary motion scenarios, endowing it with potential for widespread adoption in fields such as robotics, industrial inspection, and medical reconstruction.

\end{abstract}
\begin{IEEEkeywords}
Phase-shifting profilometry, Dynamic 3D scanning, Binomial self-compensation, Ripple-like distortions, Ghosting artifacts.
\end{IEEEkeywords}
}

\setlength{\parindent}{1em}
\maketitle
\section{Introduction}
\label{sec:intro}
\begin{figure*}[t!]
    \centering
    \includegraphics[width=0.9\linewidth]{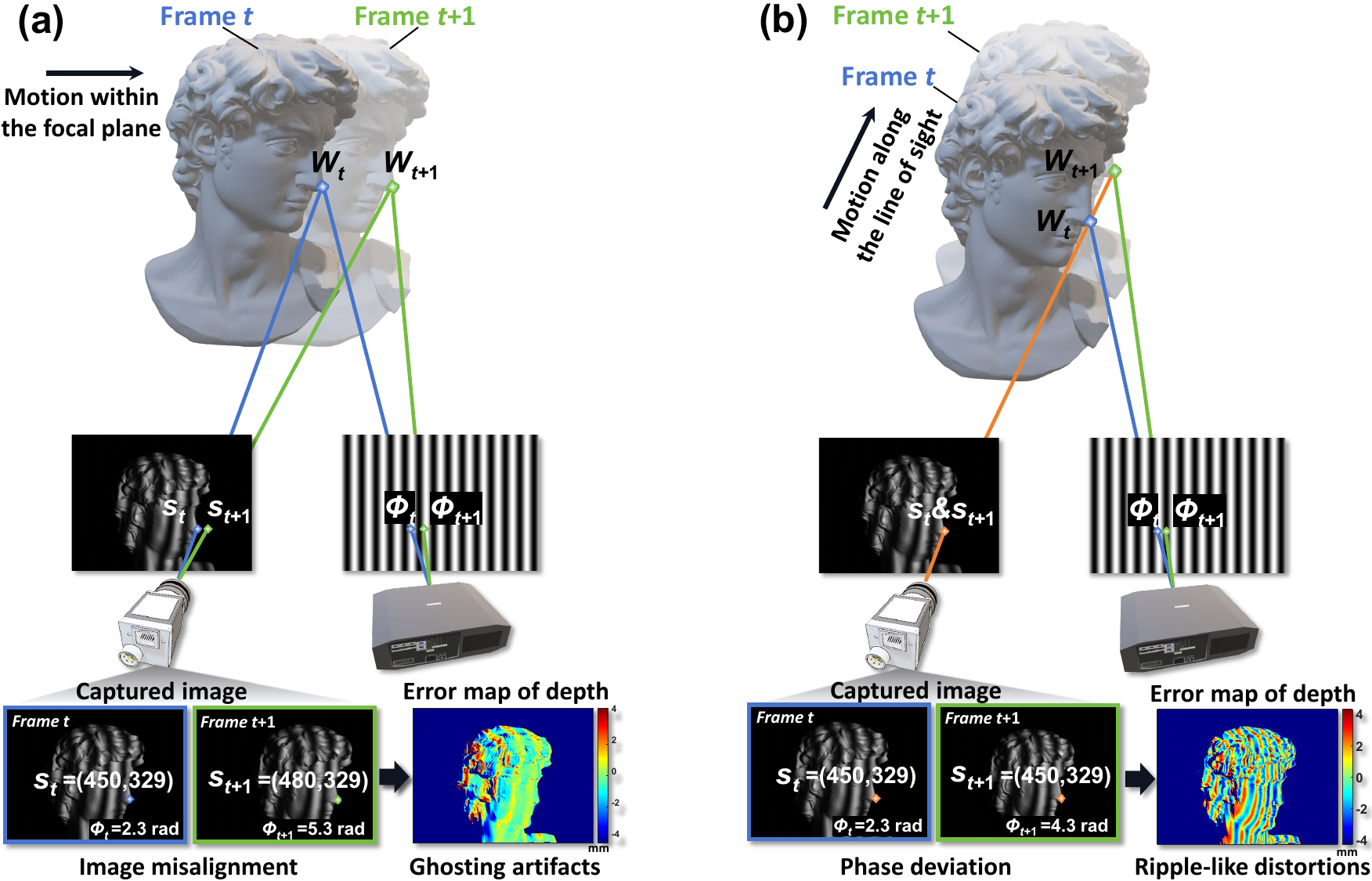}
    \caption{Diagram of motion errors in PSP: a) ghosting artifacts: induced by the image misalignment and b) ripple-like distortions: induced by the phase deviation. }
    \label{FIG:FigXYZMotion}
\end{figure*}

\IEEEPARstart{P}{hase-shifting} profilometry (PSP), known for its excellent accuracy, robustness, and pixel-wise handling~\cite{zuo2018phase}, is well-suited for high-precision 3D sensing across diverse fields such as industrial inspection~\cite{qian2021high} and design~\cite{zhu2023high}, medical reconstruction~\cite{logozzo2014recent, geng2011structured}, virtual reality~\cite{ye2020accurate}, and digital heritage preservation~\cite{wu2023high}. Research efforts have continued to explore PSP's significant capabilities across various challenging environments~\cite{wang2012robust}, covering situations with low signal-to-noise~\cite{gupta2013structured,gupta2018geometric}, multi-path interference~\cite{zhang2019causes,zhang2021sparse,zhang2022bimodalps}, high reflectivity~\cite{nayar2012diffuse}, asynchronous shooting~\cite{lu2024generative}, and global illumination effects~\cite{gupta2012combined, achar2014multi, couture2014unstructured, jiang2021parallel}, among others. However, the multi-shot nature of PSP makes it highly susceptible to object motion. This is because the fundamental assumption of PSP that the object should remain static is violated in dynamic measurement, resulting in significant motion errors in the point clouds. Although reducing the number of projected patterns can alleviate motion errors: current methods have reduced the number from 9 patterns \cite{wang2011superfast} to 5 \cite{zuo2013high}, 4 \cite{gupta2012micro}, and even fewer \cite{wu2019high, wu2019highspeed, wu2020high, zhu2022superfast}. Still, these approaches do not fundamentally resolve the issue of motion errors.

The motion error of PSP in dynamic scenes can be categorized into two main components: 1) \textbf{{ghosting artifacts}} (Fig.~\ref{FIG:FigXYZMotion}(a)): 
Focal plane motion (FPM)—motion of the object within the object-side focal plane—mainly causes misalignment among sequentially captured images, leading to ghosting artifacts~\cite{lu2013new, lu2019reconstruction, lu2023kinematic, lam2022sl,duan2024multishot} in the reconstructed depth map. These artifacts are essentially induced by image misalignment. 2) {\textbf{ripple-like distortions}} (Fig.~\ref{FIG:FigXYZMotion}(b)): Line-of-sight motion (LSM)—motion of the object along the camera's line of sight—mainly introduces sampling deviations among successive ideal phase maps, resulting in unequal phase-shifting step. These irregularities ultimately lead to ripple-like distortions~\cite{weise2007fast, liu2018motion, qian2019motion, liu2019real, wang2019motion,qian2019motion,wang2018motion,guo2021real,guo2025generalized} in the reconstructed depth map. These ripple-like distortions are essentially induced by phase deviation.

For ghosting artifacts, the image misalignment induced by the object motion constitutes the source of error. To address this issue, efforts focus on tracking the object’s trajectory. Lu et al.~\cite{lu2013new} proposed estimating the 2D rigid transformation matrix of the object by placing markers, and then calculating the motion-induced phase deviation to eliminate the error induced by in-plane motion. They further eliminate the need for placing markers on the object by adopting the speeded-up robust features algorithm~\cite{bay2006surf} to match feature points instead~\cite{lu2019reconstruction}. In recent years, they have proposed estimating the 2D rigid motion matrix by leveraging a deep learning–based optical flow tracking algorithm~\cite{lu2023kinematic}. Lam et al.\cite{lam2022sl} adopted a normalized cross-power spectrum (NCPS) to recover the object’s motion vector, from which they compensated for the phase-shifting error. Duan et al.~\cite{duan2024multishot} estimated the 2D motion vector from the motion blur using a blind image deblurring algorithm~\cite{chen2019blind}, from which they effectively mitigate both interframe ghosts and intraframe motion blur. However, object tracking-based methods are primarily specialized for 2D motion occurring perpendicularly to the sight line. Furthermore, their effectiveness is heavily dependent on the accuracy and reliability of feature tracking, which can become problematic when dealing with textureless surfaces or complex motion scenarios.

For ripple-like distortions, the primary focus is on eliminating sampling deviations. Conventional PSP relies on phase shifts that are equally distributed across a fringe period; however, this orderly distribution is disrupted by object motion. To address this, researchers estimate the motion-induced phase shift by analyzing implicit motion information embedded in the motion-affected phase frame~\cite{weise2007fast} or across multiple frames~\cite{liu2018motion, qian2019motion, liu2019real, wang2019motion}. This enables the recovery of the unknown motion-affected phase shift value for each pattern, ultimately allowing the corrected phase to be solved. Qian et al.~\cite{qian2019motion} proposed conducting a Fourier transform on each fringe pattern to have a high-frequency phase of each frame, and then estimating the motion-induced phase shifting. Wang et al.~\cite{wang2018motion} suggested using the Hilbert transform to process images and calculate the phase, which is then added to the phase obtained from the original image to automatically compensate for motion errors. Guo et al.~\cite{guo2021real} proposed summing two successive phase frames with opposite error waveforms to suppress the ripple-like distortions automatically. Further, they propose to extract the phase shift deviations from the difference information between adjacent phases~\cite{guo2025generalized}, and then an accurate phase value can be calculated. Wu et al.~\cite{wu2022suppressing} proposed a three-step gradient-based equal-step phase-shifting algorithm to effectively reduce the phase error by assuming uniform motion. Zhang et al.~\cite{zhang2024binomial} developed a phase-sequential binomial self-compensation that sums up a sequence of successive phase frames weighted by binomial coefficients, thereby effectively and flexibly eliminating the ripple-like distortions without relying on any global consistent assumption. Wang et al.~\cite{wang2025real} proposed to add a speckle pattern to the projection sequences, after which the phase deviation is computed using the digital image correlation method between the adjacent speckle patterns and then compensated. By implementing suitably chosen operators across the entire captured image~\cite{wang2018motion,wu2022suppressing} or phase frame~\cite{guo2021real}, certain methods autonomously neutralize motion-induced phase errors, producing a corrected phase directly.
 
Many existing methods primarily address either ghosting artifacts~\cite{lu2013new, lu2019reconstruction, lu2023kinematic, lam2022sl,duan2024multishot} or ripple-like distortions~\cite{weise2007fast, liu2018motion, qian2019motion, liu2019real, wang2019motion,qian2019motion,wang2018motion,guo2021real,guo2025generalized} independently, often resulting in residual inaccuracies. Recently, Li et al.~\cite{li2025mrpca} proposed a motion-resistant PSP (mr-PSP) to effectively reduce both ghosting artifacts and ripple-like distortions. Their method first applies principal component analysis to the NCPS~\cite{lam2022sl} of two successive fringe images to recover a globally consistent motion vector for image alignment. Subsequently, motion-induced phase shifts are estimated from the variance of fringe pattern differences~\cite{guo2013phase}, enabling accurate phase map computation. Although mr-PSP handles both LSM and FPM, it estimates a globally consistent motion vector and motion-induced phase shift, which limits its effectiveness in rotation and non-rigid motion scenarios. Besides Li's method \cite{li2025mrpca}, we notice that many methods assume global consistent motion. For instance, Lu et al.~\cite{lu2013new, lu2023kinematic} assumed a single global rotation matrix and translation vector. Guo et al.~\cite{guo2013phase} assumed that the phase-shifting value remains constant across the entire field. Duan et al.~\cite{duan2024multishot} employed a blind image deblurring kernel to estimate a global consistent translation vector. Similarly, Lam et al.~\cite{lam2022sl} adopted a normalized cross-power spectrum for estimating a global consistent translation vector. Wang et al.~\cite{wang2018motion} proposed using the Hilbert transform to automatically mitigate motion error, implicitly assuming a globally uniform phase shift. Since such globally consistent assumptions do not hold in non-rigid motion and multi-target scenarios, residual errors arise. As a result, existing methods are largely limited to specific motion types.

To comprehensively address both types of motion error without relying on any global consistent assumption, we propose a robust phase-shifting profilometry for arbitrary motion (RPSP-AM), including six-degrees-of-freedom (6-DoF) motion (translation and rotation in any direction), non-rigid motion, and multi-target scenarios, achieving motion-error-free 3D reconstruction. Our RPSP-AM includes optical-flow-based image alignment for ghosting artifacts, followed by image-sequential binomial self-compensation (I-BSC) for ripple-like distortions. First, we introduce two uniformly illuminated images at the beginning and end of the fringe pattern sequence to estimate the optical flow. This allows us to align the image sequence in a pixel-wise manner to eliminate the ghosting artifacts. Second, inspired by the idea of weighted sum using binomial coefficients in the P-BSC~\cite{zhang2024binomial}, we propose a high-accuracy and low-complexity I-BSC that addresses error accumulation and high computational overhead in P-BSC~\cite{zhang2024binomial}. Instead of summing \textbf{phase frames} as P-BSC~\cite{zhang2024binomial}, our I-BSC performs a weighted summation of the homogeneous \textbf{fringe images} using binomial coefficients before computing the final phase, and only requires computing the arctangent function once rather than multiple times as P-BSC. This approach exponentially reduces ripple-like distortions without introducing error accumulation. Extensive experimental results demonstrate that, under challenging conditions such as 6-DoF motion, non-rigid deformations, and multi-target movements, the proposed RPSP-AM outperforms SoTA methods in compensating for both ghosting artifacts and ripple-like distortions in a pixel-wise manner. 

The main contributions of this paper are: 
\begin{itemize}
    \item We categorize the motion errors in PSP into two components: 1) ghosting artifacts induced by image misalignment, primarily attributed to focal plane motion (FPM); 2) ripple-like distortions induced by phase deviations, primarily attributed to line-of-sight motion (LSM).
    \item For the ghosting artifacts, we design a projection strategy with two uniformly illuminated patterns at the beginning and end of the fringe pattern sequence to estimate the optical flow, from which we align the image sequence in a pixel-wise manner.    
    \item For the ripple-like distortions, we propose a high-accuracy, low-complexity image-sequential binomial self-compensation (I-BSC) method, which performs a summation of the homogeneous fringe images weighted by binomial coefficients, exponentially reducing ripple-like distortions with a competitive computational cost to that of the traditional four-step phase-shifting method.
    \item We elaborate on the error theory and computational complexity of I-BSC from the perspectives of mechanism, simulation, and experimentation. Extensive and meticulous experiments demonstrate that our method significantly suppresses motion errors under various challenging scenarios, including 6-DoF motion, non-rigid deformation, and multi-target movement, outperforming SoTA approaches.
\end{itemize}

\section{Method}
\begin{figure*}
    \centering
    \includegraphics[width=0.95\linewidth]{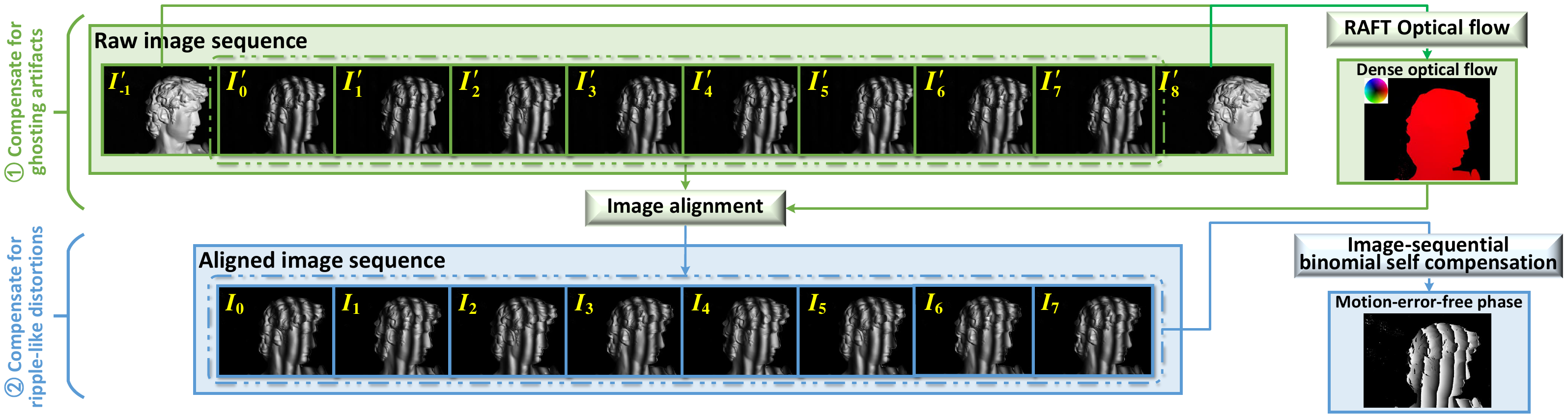}
    \caption{Pipeline of RPSP-AM. Ghosting artifacts and ripple-like distortions are eliminated successively through optical-flow-based image alignment and I-BSC. We omit the speckle pattern used for phase unwrapping in this diagram.}
    \label{FIG:FigProjectionDiagram}
\end{figure*}
In this paper, we adopt a cyclical~\cite{zhong2013fast} four-step phase-shifting strategy to project a group of high-frequency fringe patterns. In terms of phase unwrapping, we opt for a speckle-assisted stereo phase unwrapping method~\cite{deng2021high}, thereby achieving unambiguous 3D point reconstruction using a single set of high-frequency fringe patterns and one speckle pattern. Since we focus on the compensation of motion errors in this paper, further details regarding unwrapping will not be elaborated herein. In terms of nonlinearity issues, we propose a full-chain nonlinearity rectification (FNR) method in high-speed scenes, from which the nonlinear error in pattern illumination and camera digitization that emerges due to the limited number of patterns is effectively suppressed. The implementation details of FNR can be found in the supplementary document. 

When measuring a moving object using four-step phase-shifting, the captured image sequence can be expressed as
\begin{equation}
I_n'\left(\boldsymbol{c}\right) = A_n\left(\boldsymbol{c}\right) + B_n\left(\boldsymbol{c}\right) \cos\left(\phi_n\left(\boldsymbol{c}\right) \right),
\end{equation}
where $\boldsymbol{c} = (x^c, y^c)^\top$ denotes the pixel coordinate in the camera space, $A_n$ and $B_n$ represent the background intensity and the fringe modulation at frame $n$, which are mainly determined by the object texture. $\phi_n=\phi^r+\phi^h_n - \pi n/2$ is the ground-truth phase distribution at frame $n$, which consists of three components, i.e.: 1) reference phase component $\phi^r$: the phase distribution of a virtual reference plane, which remains unchanged across frames; 2) height-modulated component: the phase modulated by the height distribution of object. which changes due to the object motion across frames; 3) constant phase shifting: a constant phase shifting valure of $\pi n/2$ between two successive frames. The coordinates of corresponding points across image sequence can be expressed as $\boldsymbol{s}_n = \boldsymbol{c} + \boldsymbol{d}_n(\boldsymbol{c})$ due to the focal-plane motion (FPM), where $\boldsymbol{d}_n$ represents the displacement field between frame $0$ and frame $n$. Based on the invariance of object texture, assuming a Lambertian surface and ignoring the global illumination, we obtain two fundamental equalities during object motion: ${A_0}\left( \boldsymbol{c} \right) = {A_n}\left( \boldsymbol{s}_n \right),{B_0}\left( \boldsymbol{c} \right) = {B_n}\left( \boldsymbol{s}_n \right)$. Therefore, directly compute phase map with image sequence $I_0'\left(\boldsymbol{c}\right),I_1'\left(\boldsymbol{c}\right),...,I_n'\left(\boldsymbol{c}\right)$ introduces significant ghosting artifacts and ripple-like distortions. 
Our ultimate goal is to obtain the motion-error-free phase $\phi_0(\boldsymbol{c})$ of frame 0 by employing a two-stage scheme, which will be illustrated next to successively compensate for both errors. As illustrated in Fig.~\ref{FIG:FigProjectionDiagram}, our projection strategy employs 8 fringe patterns with a $\pi/2$ phase shift, along with uniform-intensity patterns projected at the beginning and end of the sequence.

\subsection{Eliminating Ghosting Artifacts}
\label{sec:GhostingArtifacts}
For the ghosting artifacts, once the pixel-wise displacement vector (i.e., the optical flow of the current frame) is estimated, the corresponding points across the image sequence can be aligned, thereby effectively eliminating ghosting artifacts. In this paper, we adopt RAFT network~\cite{teed2020raft} to estimate the optical flow $\boldsymbol{f}$ between frame -1 and frame 8 under uniform illumination, as illustrated in Fig.~\ref{FIG:FigProjectionDiagram}. It is worth emphasizing that any state-of-the-art optical flow estimation algorithm can be employed; RAFT is used here solely as a representative example. We propose an optical-flow-based image alignment to eliminate the ghosting artifacts. If the motion is small and linear over the image sequence, the displacement field can be estimated by $\boldsymbol{d}_n = n\boldsymbol{f}/9$, and the images without the FPM influence are obtained via looking up table: $I_n(\boldsymbol{c})=I_n'(\boldsymbol{c}+\boldsymbol{d}_n(\boldsymbol{c}))$. The aligned image sequence is
\begin{equation}   
\begin{split}
I_n\left(\boldsymbol{c}\right)&=A_n\left(\boldsymbol{s}_n\right)+B_n\left(\boldsymbol{s}_n\right)\cos\left(\phi_n\left(\boldsymbol{s}_n\right)\right)\\
&=A_0\left(\boldsymbol{c}\right)+B_0\left(\boldsymbol{c}\right)\cos\left(\phi_n\left(\boldsymbol{c}\right)+x_n\left(\boldsymbol{c}\right)\right),
\end{split}
\label{EQ:LSM}
\end{equation}
where
\begin{equation}   
x_n\left(\boldsymbol{c}\right)=\alpha\left(\boldsymbol{s}_n\left(\boldsymbol{c}\right)\right)+\beta_n\left(\boldsymbol{s}_n\left(\boldsymbol{c}\right)\right)
\nonumber
\end{equation}
with
\begin{equation}  
\left\{\begin{split}
\alpha\left(\boldsymbol{s}_n\left(\boldsymbol{c}\right)\right)={\phi ^r}\left( {{\boldsymbol{s}_n}\left(\boldsymbol{c}\right)} \right) - {\phi ^r}\left( {{\boldsymbol{c}}} \right)\\
\beta_n\left(\boldsymbol{s}_n\left(\boldsymbol{c}\right)\right)=\phi _n^h\left( {{\boldsymbol{s}_n\left(\boldsymbol{c}\right)}} \right) - \phi_n ^h\left( {{\boldsymbol{c}}} \right)
\end{split}\right.,
\end{equation}
where $x_n$ denotes the total phase deviations, $\alpha$ and $\beta_n$ denote the phase deviations induced by FPM and LSM in the aligned image sequence, respectively. Note that $\beta_n$ inherently exists because of the movement of the object, while $\alpha$ is introduced during the image alignment. 

After image alignment, the phase is computed as 
\begin{equation} 
\tilde \phi_n  = {\tan ^{ - 1}}\left[ {\frac{{{I_{n+1}} - {I}_{n+3}}}{{{I_{n}} - {I}_{n+2}}}} \right].
\label{EQ:FourStepPhase}
\end{equation}
By substituting Eq.~(\ref{EQ:LSM}) into (\ref{EQ:FourStepPhase}) and applying the approximation $\tan(\cdot)\approx\cdot$, we derive the expression of the ripple-like distortions and omit the index $\boldsymbol{c}$ as
\begin{equation}
    \label{EQ:MotionError4}
\begin{split}
     &{\epsilon_{\text{ripple}}}\left( \tilde{{\phi}}_n \right)= \tilde{\phi_n} - \phi _n\\
    &\approx     \frac{1}{4}\sum\limits_{i = 0}^3 {\left[ {{x_{n+i}} - {{\left( { - 1} \right)}^{n+i}}{x_{n+i}}\cos \left( {2\phi_0}\right)}\right]}
\end{split}.
\end{equation}
It is evident that the error formula is free of any terms related to the object textures $A_0$ and $B_0$, indicating that the ghosting artifacts have been effectively eliminated. The residual motion error comprises a DC component and a harmonic component with twice the frequency of the wrapped phase. The DC component, manifesting as the lag of the point clouds due to object motion, is independent of the phase value, and thus is irrelevant to the ripple-shaped error. Consequently, the crux of motion error compensation lies in eliminating the harmonic. Therefore, the ripple-like distortions still persists after image alignment, necessitating further processing in the next section. Correcting for ripple-like distortions is critical not only because pure LSM significantly degrades PSP accuracy in dynamic scenes, but also because image alignment during correction of ghosting artifacts can introduce potential phase sampling deviations similar to those caused by LSM.

\subsection{Compensating for Ripple-like Distortions}
\begin{figure*}[t!]
    \centering
    \includegraphics[width=0.85\linewidth]{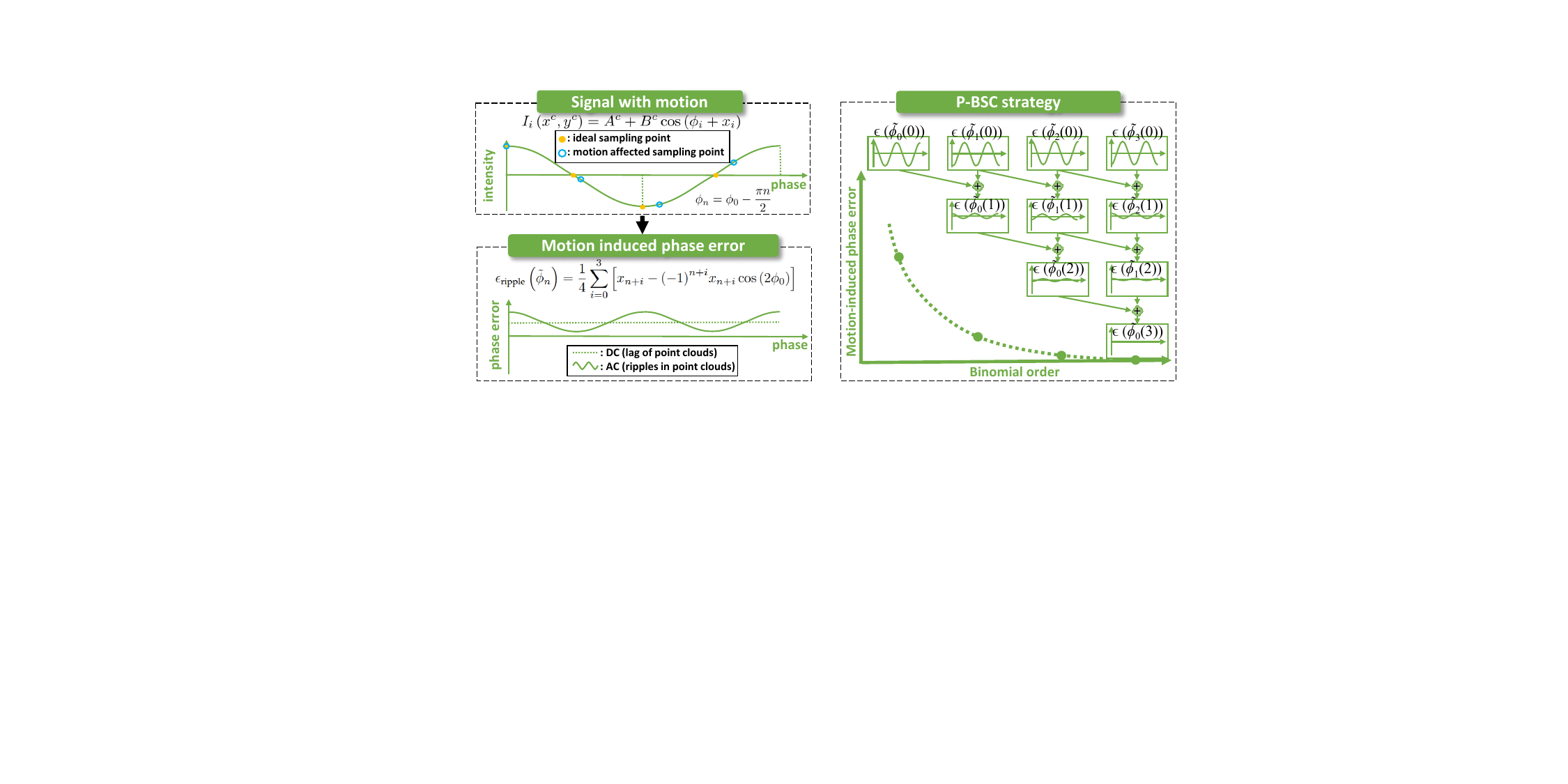}
    \caption{Left: mechanism and characteristic of ripple-like distortions. Right: P-BSC~\cite{zhang2024binomial} exponentially reduces the motion error by leveraging the inherent property of the phase sequence. Inspired by P-BSC, we propose an I-BSC to more effectively and efficiently suppress ripple-like distortions by directly processing the image sequence.}
    \label{FIG:FigMotionErrorMechanism}
\end{figure*}

For the ripple-like distortions, before introducing I-BSC, we briefly explain the error suppression mechanism of P-BSC~\cite{zhang2024binomial}, which inspired our approach. By analyzing the motion error waveform of two successive phase frames $\tilde \phi _n$ and $\tilde \phi _{n+1}$ in a discrete time series, we discover that the coefficients of trigonometric terms exhibit opposite signs owing to a phase shift of $\pi/2$ as shown in Fig.~\ref{FIG:FigMotionErrorMechanism}. Consequently, P-BSC~\cite{zhang2024binomial} recommonded to sum each pair of successive phase frames, and sum the resultant phase frames again and again until there is only one frame left. Such operation is equivalent to summing up the phase sequence weighted by $K$-th order binomial coefficients~\cite{zhang2024binomial} 
\begin{equation}\label{EQ:YangHuiSum}
\tilde{\phi}_n^K = {2^{ - K}}\sum\limits_{k = 0}^K {\binom{K}{k}\tilde \phi _{n + K - k}}.
\end{equation}

Considering $x_n$ and its high order difference
\begin{equation}
\label{EQ:HighOrderDifference}
\left\{ \begin{split}
{\Delta ^{\left( K \right)}}{x_n} = {\Delta ^{\left( {K - 1} \right)}}{x_{n + 1}} - {\Delta ^{\left( {K - 1} \right)}}{x_n}\\
{\Delta ^{\left( K \right)}}{x_n} = \sum\limits_{k = 0}^K {{{( - 1)}^k}} \binom{K}{k}{x_{n + K - k}}
\end{split} \right.,
\end{equation}
where we define ${\Delta ^{\left( 0 \right)}}{x_i}=x_i$, we substitute Eq.~(\ref{EQ:MotionError4}) into Eq.~(\ref{EQ:YangHuiSum}) and obtain the result in~\cite{zhang2024binomial}: 
\begin{equation}
\label{EQ:YangHuiSum4}
\tilde{\phi}_n^K = \bar\phi _n^K  + \underbrace{{2^{ - \left( {K + 2} \right)}}\left( {{\Delta ^{\left( {K + 1} \right)}}{x_n} + {\Delta ^{\left( {K + 1} \right)}}{x_{n + 2}}} \right)\cos \left( {2{\phi _0}} \right)}_{\epsilon_{\text{P-BSC}}},
\end{equation}
where 
\begin{equation}
\bar \phi _n^{K} = {2^{ - K}}\sum\limits_{k = 0}^K {\binom{K}{k}\left( {{\phi _{n + k}} + \frac{1}{4}\sum\limits_{i = 0}^3 {{x_{n + k + i}}} } \right)}.
\end{equation}
Based on Eq.~(\ref{EQ:YangHuiSum4}), it is clear that increasing $K$ produces two main effects: first, the factor $2^{-(K+2)}$ exponentially decreases the harmonic amplitudes; second, higher-order difference terms (approach zero as $K$ increases) occur in the harmonic amplitudes. Both effects significantly diminish the ripple-like distortions as depicted in Fig.~\ref{FIG:FigMotionErrorMechanism}.

\subsubsection{Image-Sequential Binomial Self-Compensation}
\begin{figure*}[ht!]
    \centering
    \includegraphics[width=0.95\linewidth]{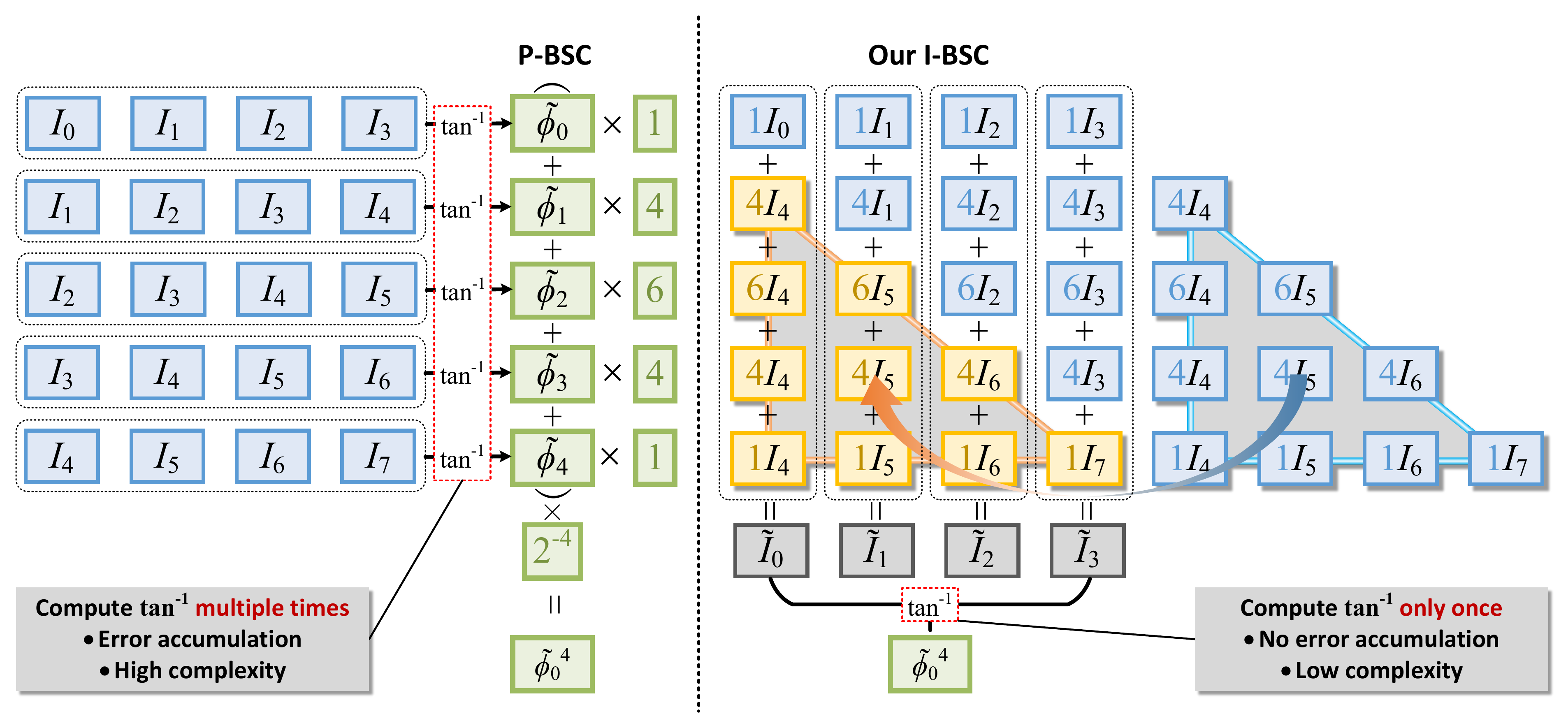}
    \caption{Comparison between P-BSC~\cite{zhang2024binomial} and our I-BSC. P-BSC~\cite{zhang2024binomial} computes the arctangent function $K+1$ times to have a phase sequence, and then sums the phase sequence weighted by binomial coefficients. 
    Our I-BSC sums the homogeneous fringe images, weighted by binomial coefficients, to generate four compensated fringe images, and then compute the motion-error-free phase. The arctangent function is computed only once. }
    \label{FIG:FigIBSC}
\end{figure*}

Although P-BSC~\cite{zhang2024binomial} effectively suppresses ripple-like distortions, it faces two challenges: high computational overhead and error accumulation, arising from the need to calculate multi-frame phases and perform weighted summation. In this paper, inspired by P-BSC, we propose a novel I-BSC that computes the arctangent function only once, addressing both challenges. Our I-BSC is inspired by idea of summing up the phase sequence weighted by binomial coefficients in P-BSC. Instead of summing up the \textbf{phase frames}, our I-BSC adds up the homogeneous \textbf{fringe images} weighted by the binomial coefficients, from which four compensated fringe images $\tilde{I}_0,\tilde{I}_1,\tilde{I}_2,\tilde{I}_3$ are obtained as shown in Fig.~\ref{FIG:FigIBSC}(a). The homogeneous fringe images mean the encoded frames with the same trigonometric function name and polarity. For example, $I_{4n},I_{4n+1},I_{4n+2},I_{4n+3}$ are four groups of homogeneous fringe images, where $n\in\mathbb{N}$. The four compensated fringe images are computed as
\begin{equation}
{\tilde I_m} = \sum\limits_{k = 0}^K {\binom{K}{k}} {I_{{V_m}\left( k \right)}},  
\label{EQ:CompensatedImageFrames}
\end{equation}
where the index vector is
\begin{equation}
{V_m}\left( k \right) = \left( {k + 3} \right) - \left[ {\left( {k + 3 - m} \right)\bmod \left( 4 \right)} \right],
\end{equation}
where $m\in0,1,2,3$ represents the group number of homogeneous fringe images. Finally, only one arctangent function is employed to compute the motion-error-free phase as
\begin{equation}
{{\tilde \phi }_0^ K} = {\tan ^{ - 1}}\left( {\frac{{{{\tilde I}_1} - {{\tilde I}_3}}}{{{{\tilde I}_0} - {{\tilde I}_2}}}} \right).
\label{EQ:IBSC}
\end{equation}
An additional advantage of using I-BSC is that the ripple-error-free modulation image $B$ can be facilely computed as
\begin{equation}
B = {2^{ - \left( {K + 1} \right)}}\sqrt {{{\left( {{{\tilde I}_1} - {{\tilde I}_3}} \right)}^2} + {{\left( {{{\tilde I}_0} - {{\tilde I}_2}} \right)}^2}},
\end{equation}
while P-BSC does not provide a method for calculating $B$.

The I-BSC framework is concise and can be described using just two equations (i.e., Eq.~(\ref{EQ:CompensatedImageFrames}) and (\ref{EQ:IBSC})). 
The I-BSC algorithm is conceptually inspired by the intuitive idea of applying binomial coefficients as summation weight in P-BSC~\cite{zhang2024binomial}, rather than being directly derived from its mathematical formulation. Specifically, in our I-BSC framework, rather than assigning binomial coefficients to each phase frame, we reassign them to the set of four constituent fringe images required for computing the corresponding phase frame, as depicted in Fig.~\ref{FIG:FigIBSC}(a), from which we generalize the BSC concept from phase sequences to image sequences. As shown in the following analysis, the theoretical error of I-BSC is identical to that of P-BSC. However, in practice, it is free from error accumulation, resulting in higher accuracy. Meanwhile, the required number of arctangent function calculations is reduced from $K$ to 1, significantly improving the computation speed. 

\begin{figure}[h!]
    \centering
    \includegraphics[width=0.85\linewidth]{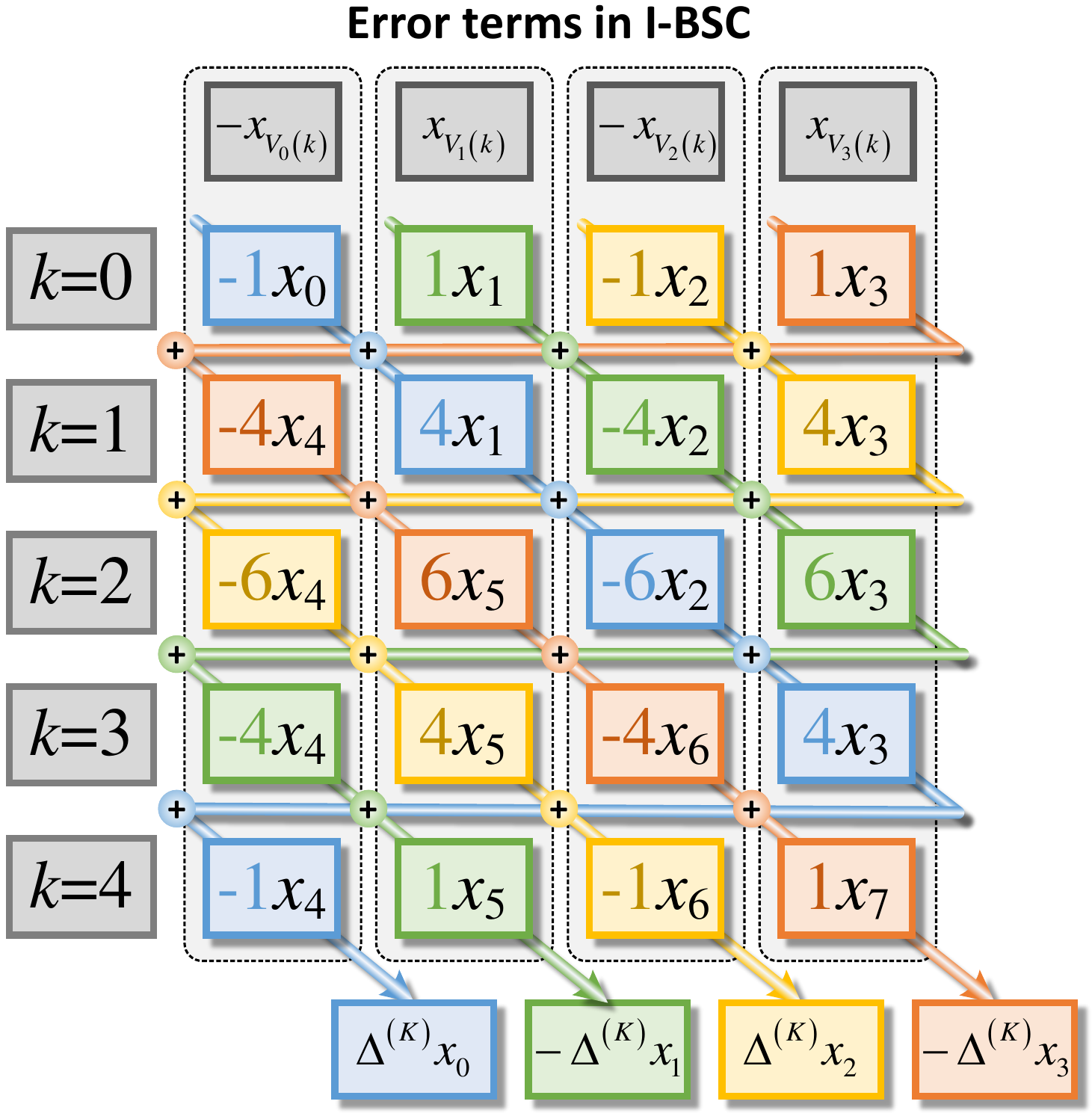}
    \caption{The sum of the error terms along the diagonal direction precisely yields the high-order difference form error terms (Eq.~(\ref{EQ:IBSCError}) to Eq.~(\ref{EQ:IBSCError2}))..}
    \label{FIG:FigIBSCError}
\end{figure}

\vspace{1em}
\begin{strip}
To model the error of I-BSC, considering a small phase-shift error $x_{V_m(k)}$, we have $\sin(x_{V_m(k)})\approx x_{V_m(k)}$ and $\cos(x_{V_m(k)})\approx 1$. Subsequently, we rewrite Eq.~(\ref{EQ:LSM}) as 
\begin{equation}
{I_{{V_m}\left( k \right)}} = A_0 + B_0\cos \left( {{\phi _{{V_m}\left( k \right)}}} \right) - B_0{x_{{V_m}\left( k \right)}}\sin \left( {{\phi _{{V_m}\left( k \right)}}} \right).
\label{EQ:CapturedImage2}
\end{equation}
Since there are numerous index variables involved in this section, we will focus on the computation of the phase for frame 0. The calculation and analysis for frame $n$ follow a similar approach. Then we substitute Eq.~(\ref{EQ:CapturedImage2}) into Eq.~(\ref{EQ:IBSC}) to have
\begin{equation}
\begin{split}
{{\tilde \phi }_0^K}&= {\tan ^{ - 1}}\left( {\frac{{\sum\limits_{k = 0}^K {\binom{K}{k}} \left( {{I_{{V_3}\left( k \right)}} - {I_{{V_1}\left( k \right)}}} \right)}}{{\sum\limits_{k = 0}^K {\binom{K}{k}} \left( {{I_{{V_2}\left( k \right)}} - {I_{{V_0}\left( k \right)}}} \right)}}} \right)\\
 &= {\tan ^{ - 1}}\left( {\frac{{\sin \left( {{\phi _0}} \right) + {2^{ - \left( {K + 1} \right)}}\sum\limits_{k = 0}^K {\binom{K}{k}\left( {{x_{{V_3}\left( k \right)}} + {x_{{V_1}\left( k \right)}}} \right)\cos \left( {{\phi _0}} \right)} }}{{\cos \left( {{\phi _0}} \right) - {2^{ - \left( {K + 1} \right)}}\sum\limits_{k = 0}^K {\binom{K}{k}\left( {{x_{{V_2}\left( k \right)}} + {x_{{V_0}\left( k \right)}}} \right)\sin \left( {{\phi _0}} \right)} }}} \right)
\end{split}.
\end{equation}
Thus, the motion error of I-BSC is expressed as
\begin{equation}
\begin{split}
{\epsilon_{{\rm{I - BSC}}}} &= {{\tilde \phi }_0^K} - {\phi _0}\\
 &= {\tan ^{ - 1}}\left( {\frac{{\sum\limits_{k = 0}^K {\binom{K}{k}\left( {{x_{{V_3}\left( k \right)}} + {x_{{V_1}\left( k \right)}}} \right){{\cos }^2}\left( {{\phi _0}} \right)}  + \sum\limits_{k = 0}^K {\binom{K}{k}\left( {{x_{{V_2}\left( k \right)}} + {x_{{V_0}\left( k \right)}}} \right){{\sin }^2}\left( {{\phi _0}} \right)} }}{{{2^{K + 1}} + \sum\limits_{k = 0}^K {\binom{K}{k}\left[ {\left( {{x_{{V_3}\left( k \right)}} + {x_{{V_1}\left( k \right)}}} \right) - \left( {{x_{{V_2}\left( k \right)}} + {x_{{V_0}\left( k \right)}}} \right)} \right]\sin \left( {{\phi _0}} \right)\cos \left( {{\phi _0}} \right)} }}} \right)\\
 &\approx {2^{ - \left( {K + 2} \right)}}\sum\limits_{k = 0}^K {\binom{K}{k}\left[ {\sum\limits_{m = 0}^3 {{x_{{V_m}\left( k \right)}}}  + \left( { - {x_{{V_0}\left( k \right)}} + {x_{{V_1}\left( k \right)}} - {x_{{V_2}\left( k \right)}} + {x_{{V_3}\left( k \right)}}} \right)\cos \left( {2{\phi _0}} \right)} \right]} \\
& := {2^{ - \left( {K + 2} \right)}}\sum\limits_{k = 0}^K {\binom{K}{k}\left( { - {x_{{V_0}\left( k \right)}} + {x_{{V_1}\left( k \right)}} - {x_{{V_2}\left( k \right)}} + {x_{{V_3}\left( k \right)}}} \right)\cos \left( {2{\phi _0}} \right)} 
\end{split},
\label{EQ:IBSCError}
\end{equation}
The DC component of the error is omitted because it does not significantly affect the reconstruction results. Analytical derivation of a concise error expression from Eq.~(\ref{EQ:IBSCError}) is not straightforward. For a more intuitive view, Fig.~\ref{FIG:FigIBSCError} lists the error terms in I-BSC for $k=0\sim4$. We find that neither row-wise nor column-wise summation of the error terms yields a simplified expression. Fortunately, the relation $V_{(m+1)\text{mod}(4)}(k+1)\equiv V_{m}(k)+1$ is observed. This implies that along the diagonal direction, with both the row index $k$ and the column index $m$ increasing by one, the subscript of the error term $x$ increments by one while its sign alternates, as depicted in Fig.~\ref{FIG:FigIBSCError}. Therefore, summing the terms along the diagonal direction rather than row-by-row or column-by-column, in light of Eq.~(\ref{EQ:HighOrderDifference}), allows for the immediate derivation of the high-order difference form error terms. Ultimately, Eq.~(\ref{EQ:IBSCError}) is simplified into
\begin{equation}
\begin{split}
{\epsilon_{{\rm{I - BSC}}}} &= {2^{ - \left( {K + 2} \right)}}\sum\limits_{k = 0}^K {{{\left( { - 1} \right)}^k}\binom{K}{k}\left( { - {x_k} + {x_{k + 1}} - {x_{k + 2}} + {x_{k + 3}}} \right)}\cos \left( {2{\phi _0}} \right) \\
 &= {2^{ - \left( {K + 2} \right)}}{\left( { - 1} \right)^K}\sum\limits_{k = 0}^K {{{\left( { - 1} \right)}^{K - k}}\binom{K}{k}\left( { - {x_k} + {x_{k + 1}} - {x_{k + 2}} + {x_{k + 3}}} \right)}\cos \left( {2{\phi _0}} \right) \\
 &= {2^{ - \left( {K + 2} \right)}}{\left( { - 1} \right)^K}\left( {{\Delta ^{\left( {K + 1} \right)}}{x_0} + {\Delta ^{\left( {K + 1} \right)}}{x_2}} \right)\cos \left( {2{\phi _0}} \right)
\end{split}.
\label{EQ:IBSCError2}
\end{equation}
\end{strip}

A comparison of the theoretical expressions in Eq. (\ref{EQ:YangHuiSum4}) and Eq. (\ref{EQ:IBSCError2}) indicates that I-BSC and P-BSC exhibit the same theoretical performance in motion error suppression. However, in practice, I-BSC does not sum phase frames up, thereby avoiding error accumulation induced by approximation of $\tan(\cdot)\approx\cdot$ in Eq.~(\ref{EQ:YangHuiSum4}), whereas P-BSC retains the residual errors. As a result, I-BSC converges faster than P-BSC as $K$ increases, which will be verified and elaborated on in simulations and real-world experiments in Section~\ref{sectionConvergenceSpeed} and \ref{SEC:ComputationalEfficiency}. The I-BSC procedure is summarized as Algorithm~\ref{ALG:IBSC}. The MATLAB code for implementing I-BSC is open sourced at \url{https://github.com/GeyouZhang/I-BSC}.

\begin{algorithm}
    \small
    \caption{I-BSC for motion error} 
    \label{ALG:IBSC}
    \KwIn
    {			        
        \\Binomial order: $K$.				
        \\Captured image sequence: $\left\{I_0,...,I_{K+4}\right\}$.
    }
    \KwOut
    {
        \\Motion-error-free phase: $\tilde\phi_{0}^K$.			
    }	
    \For{ $m \leftarrow 0$ \KwTo $3$ }	
    {	
        $\tilde{I}_m\leftarrow0$\\
        \For{ $k \leftarrow 0$ \KwTo $K$ }	
        {	
        ${V_m}\left( k \right) \leftarrow \left( {k + 3} \right) - \left[ {\left( {k + 3 - m} \right)\bmod \left( 4 \right)} \right]$\\
        $\tilde{I}_m\leftarrow\tilde{I}_m+\binom{K}{k}I_{{V_m}\left( k \right)}$
        }
    }    
    $\tilde\phi_{0}^K\leftarrow {\tan ^{ - 1}}\left( {\frac{{{{\tilde I}_1} - {{\tilde I}_3}}}{{{{\tilde I}_0} - {{\tilde I}_2}}}} \right) $ 
\end{algorithm}


\section{Experimental Evaluations}
\label{SEC:Exp}
The proposed algorithm is implemented on an Intel(R) Core(TM) i7-13790F @ 2.10 GHz with 32 GB RAM. Our experimental system consists of two AVT 1800U-052 monochrome cameras that have $816\times624$ resolution, equipped with 8~mm Computar lenses, and a TI DLP4500 projector with a resolution of $912\times1140$, from which we can conduct speckle-assisted stereo phase unwrapping. The two cameras are synchronized by the trigger signal from the projector.
\subsection{Quantitative Evaluation}
\label{sectionA}
\begin{figure}[t!]
    \centering
    \includegraphics[width=1\linewidth]{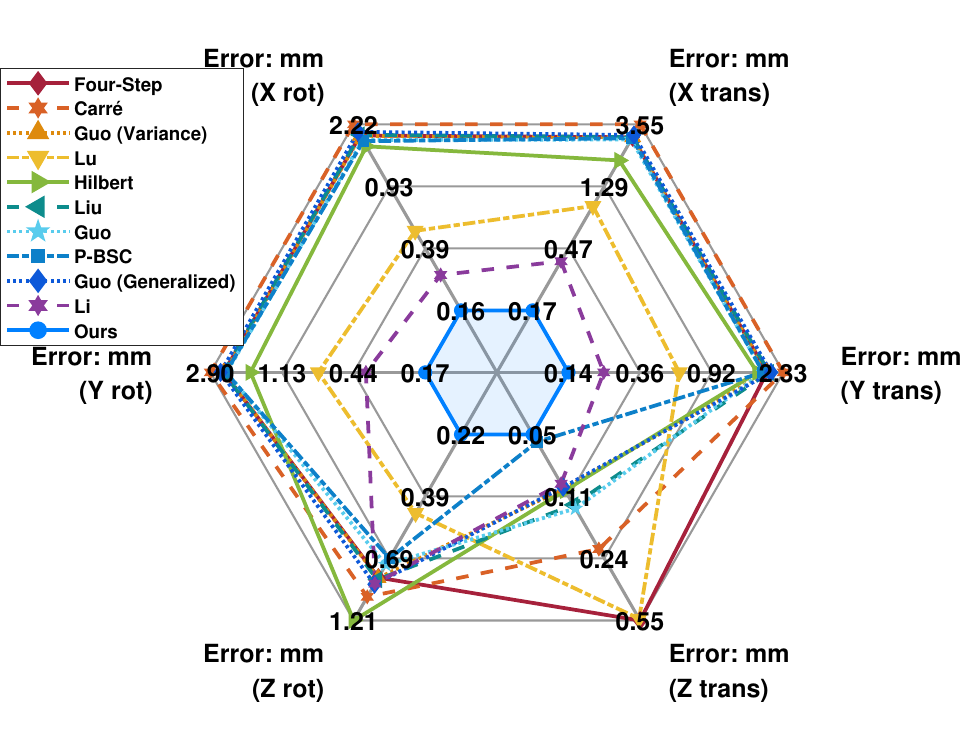}  
    \caption{ Spider plot of RMSE of our RPSP-AM and comparison methods under 6-DoF motion (including translation and rotation in any direction). Comparison methods include: 1) traditional four-step PSP, 2) Carré's~\cite{carre1966installation}, 3) Guo's~\cite{guo2013phase} (variance), 4) Lu's~\cite{lu2013new}, 5) Wang's~\cite{wang2018motion}, 6) Liu's~\cite{liu2019real}, 7) Guo's~\cite{guo2021real}, 8) our previous proposed P-BSC~\cite{zhang2024binomial}, 9) Guo's~\cite{guo2025generalized} (generalized), and 10) Li's~\cite{li2025mrpca}. Our method showcases superior performance in all six motion scenarios, significantly suppressing motion error.} 
    \label{FIG:FigSpeedVSRMSESpider}
\end{figure}

\begin{figure*}[t!]
    \centering
    \includegraphics[width=1\linewidth]{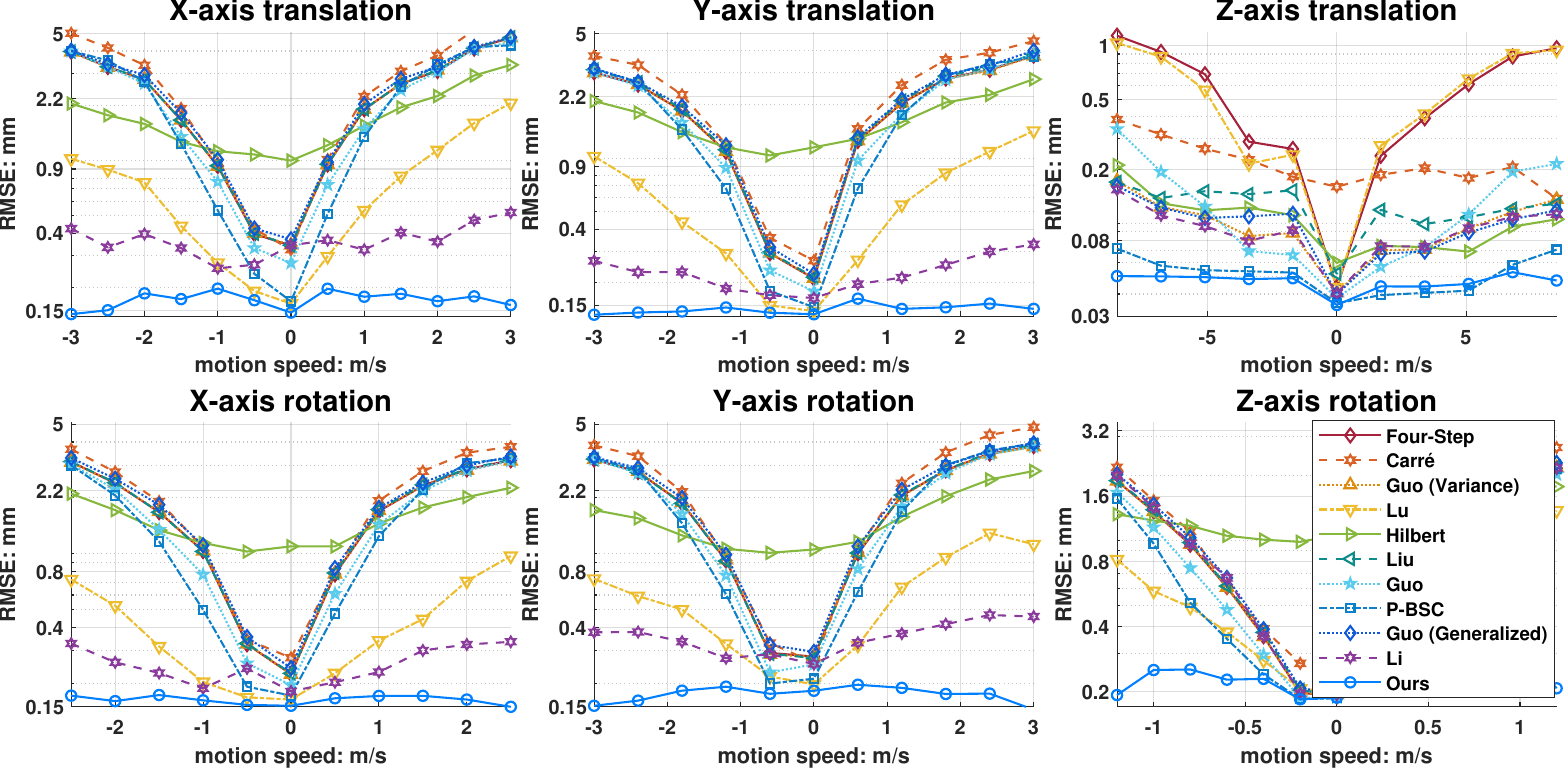}  
    \caption{Measurement error VS motion speed under 6-DoF motion (including translation and rotation in any direction). We measured a uniform white standard plate translating along the Z-axis and a standard flat plate with an array of circular markers translating along X- and Y-axes and rotating around X-, Y-, and Z-axes. Comparison methods include: 1) traditional four-step PSP~(4 frames), 2) Carré's~\cite{carre1966installation}~(4 frmaes), 3) Guo's~\cite{guo2013phase} phase shift estimation from variances~(4 frmaes), 4) Lu's~\cite{lu2013new} method (4 frames), 5) Wang's~\cite{wang2018motion} Hilbert transform compensation~(4 frames), 6) Liu's~\cite{liu2019real} method~(8 frames), 7) Guo's~\cite{guo2021real} phase frame sum method~(5 frmaes), 8) our previous proposed P-BSC~\cite{zhang2024binomial}~(8 frames), 9) Guo's~\cite{guo2025generalized} generalized phase shift deviation estimation~(6 frames), 10) Li's~\cite{li2025mrpca} motion-resistant PSP~(4 frmaes), and 11) our RPSP-AM~(8 fringes and 2 uniform patterns). The vertical axis is on a logarithmic scale. Our BSC surpasses existing methods in compensating for the motion error under all six motion scenarios. This indicates our method's comprehensive capability to thoroughly and significantly suppress motion errors, including both ripple-like distortions and ghosting artifacts, across arbitrary motion scenarios. } 
    \label{FIG:FigSpeedVSRMSE}
\end{figure*}

\begin{table*}[ht!]    
\centering
\caption{RMSE of different methods under 6-DoF motion. The ratios to the errors of conventional four-step phase-shifting are labeled in parentheses. For each type of motion, the method with the lowest error is bolded, and the second-lowest is underlined. Quantitative assessments indicate that under arbitrary 6-DoF motion conditions, our RPSP-AM exhibits superior performance, while existing methods are typically only effective for specific types of motion. (unit: mm)}
\renewcommand{\arraystretch}{1.2}
\begin{tabular}{|c|c|c|c|c|c|c|c|c}
\hline
Method     & \multicolumn{1}{c|}{\textbf{X translation}} & \multicolumn{1}{c|}{\textbf{Y translation}} & \multicolumn{1}{c|}{\textbf{Z translation}} & \multicolumn{1}{c|}{\textbf{X rotation}} & \multicolumn{1}{c|}{\textbf{Y rotation}} & \multicolumn{1}{c|}{\textbf{Z rotation}} & \multicolumn{1}{c|}{\textbf{Average}} \\
\hline
\textbf{Four-Step} & 2.871 (1.00×) & 1.878 (1.00×) & 0.551 (1.00×) & 1.901 (1.00×) & 2.434 (1.00×) & 0.818 (1.00×) & 1.742 (1.00×) \\
\textbf{Carré~\cite{carre1966installation}} & 3.546 (1.24×) & 2.326 (1.24×) & 0.213 (0.39×) & 2.224 (1.17×) & 2.904 (1.19×) & 0.975 (1.19×) & 2.031 (1.17×) \\
\textbf{Guo (Variance)~\cite{guo2013phase}} & 2.879 (1.00×) & 1.871 (1.00×) & 0.095 (0.17×) & 1.892 (1.00×) & 2.429 (1.00×) & 0.821 (1.00×) & 1.665 (0.96×) \\
\textbf{Lu~\cite{lu2013new}} & 0.937 (0.33×) & 0.600 (0.32×) & 0.538 (0.98×) & 0.498 (0.26×) & 0.701 (0.29×) & \underline{0.454 (0.56×)} & 0.621 (0.36×) \\
\textbf{Wang~\cite{wang2018motion}} & 1.970 (0.69×) & 1.658 (0.88×) & 0.098 (0.18×) & 1.633 (0.86×) & 1.713 (0.70×) & 1.209 (1.48×) & 1.380 (0.79×) \\
\textbf{Liu~\cite{liu2019real}} & 2.880 (1.00×) & 1.931 (1.03×) & 0.118 (0.21×) & 1.918 (1.01×) & 2.448 (1.01×) & 0.833 (1.02×) & 1.688 (0.97×) \\
\textbf{Guo~\cite{guo2021real}} & 2.768 (0.96×) & 1.788 (0.95×) & 0.124 (0.22×) & 1.784 (0.94×) & 2.348 (0.96×) & 0.727 (0.89×) & 1.590 (0.91×) \\
\textbf{P-BSC~\cite{zhang2024binomial}} & 2.829 (0.99×) & 1.788 (0.95×) & \underline{0.051 (0.09×)} & 1.755 (0.92×) & 2.373 (0.98×) & 0.687 (0.84×) & 1.580 (0.91×) \\
\textbf{Guo (Generalized)~\cite{guo2025generalized}} & 2.979 (1.04×) & 1.982 (1.06×) & 0.093 (0.17×) & 2.000 (1.05×) & 2.540 (1.04×) & 0.869 (1.06×) & 1.744 (1.00×) \\
\textbf{Li~\cite{li2025mrpca}} & \underline{0.381 (0.13×)} & \underline{0.224 (0.12×)} & 0.089 (0.16×) & \underline{0.268 (0.14×)} & \underline{0.375 (0.15×)} & 0.864 (1.06×) & \underline{0.367 (0.21×)} \\
\textbf{Ours} & \textbf{0.171 (0.06×)} & \textbf{0.142 (0.08×)} & \textbf{0.046 (0.08×)} & \textbf{0.163 (0.09×)} & \textbf{0.171 (0.07×)} & \textbf{0.222 (0.27×)} & \textbf{0.153 (0.09×)}\\
\hline
\end{tabular}
\label{TAB:RMSE}
\end{table*}

\begin{table*}[ht!]    
\centering
\caption{RMSE of different methods under 6-DoF motion, all comparison methods are enhanced using optical-flow-based image alignment. "\textit{OF}" represents optical flow enhanced version of the original method. The ratios to the errors of conventional four-step phase-shifting are labeled in parentheses. For each type of motion, the method with the lowest error is bolded, and the second-lowest is underlined. Results show that under arbitrary 6-DoF motion conditions, even if all the comparison methods are enhanced using optical-flow-based image alignment, our RPSP-AM still exhibits superior performance. (unit: mm)}
\renewcommand{\arraystretch}{1.2}
\begin{tabular}{|c|c|c|c|c|c|c|c|c}
\hline
Method     & \multicolumn{1}{c|}{\textbf{X translation}} & \multicolumn{1}{c|}{\textbf{Y translation}} & \multicolumn{1}{c|}{\textbf{Z translation}} & \multicolumn{1}{c|}{\textbf{X rotation}} & \multicolumn{1}{c|}{\textbf{Y rotation}} & \multicolumn{1}{c|}{\textbf{Z rotation}} & \multicolumn{1}{c|}{\textbf{Average}} \\
\hline
\textbf{Four-Step}-\textit{OF} & 0.672 (1.00×) & 0.156 (1.00×) & 0.467 (1.00×) & 0.197 (1.00×) & 0.815 (1.00×) & 0.266 (1.00×) & 0.429 (1.00×)\\
\textbf{Carré~\cite{carre1966installation}}-\textit{OF} & 0.317 (0.47×) & 0.245 (1.57×) & 0.199 (0.43×) & 0.254 (1.29×) & 0.297 (0.36×) & 0.292 (1.10×) & 0.267 (0.62×) \\
\textbf{Guo (Variance)~\cite{guo2013phase}} \textit{OF} & 0.206 (0.31×) & 0.151 (0.97×) & 0.082 (0.18×) & 0.189 (0.96×) & 0.211 (0.26×) & 0.260 (0.98×) & 0.183 (0.43×) \\
\textbf{Lu~\cite{lu2013new}} \textit{OF} & 0.201 (0.30×) & 0.150 (0.97×) & 0.533 (1.14×) & 0.189 (0.96×) & 0.217 (0.27×) & \underline{0.230 (0.87×)} & 0.253 (0.59×) \\
\textbf{Wang~\cite{wang2018motion}} \textit{OF} & 1.141 (1.70×) & 0.998 (6.40×) & 0.090 (0.19×) & 1.013 (5.14×) & 1.009 (1.24×) & 1.032 (3.88×) & 0.881 (2.05×) \\
\textbf{Liu~\cite{liu2019real}} \textit{OF} & 0.199 (0.30×) & 0.156 (1.00×) & 0.123 (0.26×) & 0.189 (0.96×) & 0.213 (0.26×) & 0.242 (0.91×) & 0.187 (0.44×) \\
\textbf{Guo~\cite{guo2021real}} \textit{OF} & 0.224 (0.33×) & 0.145 (0.93×) & 0.104 (0.22×) & 0.172 (0.87×) & 0.243 (0.30×) & 0.231 (0.87×) & 0.186 (0.43×) \\
\textbf{P-BSC~\cite{zhang2024binomial}} \textit{OF} & \underline{0.181 (0.27×)} & \underline{0.144 (0.92×)} & \underline{0.054 (0.11×)} & \underline{0.166 (0.84×)} & \underline{0.181 (0.22×)} & 0.231 (0.87×) & \underline{0.159 (0.37×)} \\
\textbf{Guo (Generalized)~\cite{guo2025generalized}} \textit{OF} & 0.209 (0.31×) & 0.157 (1.01×) & 0.081 (0.17×) & 0.199 (1.01×) & 0.220 (0.27×) & 0.241 (0.91×) & 0.184 (0.43×) \\
\textbf{Li~\cite{li2025mrpca}} \textit{OF} & 0.206 (0.31×) & 0.150 (0.97×) & 0.082 (0.18×) & 0.189 (0.96×) & 0.211 (0.26×) & 0.259 (0.97×) & 0.183 (0.43×) \\
\textbf{Ours} & \textbf{0.171 (0.25×)} & \textbf{0.142 (0.91×)} & \textbf{0.046 (0.10×)} & \textbf{0.163 (0.83×)} & \textbf{0.171 (0.21×)} & \textbf{0.222 (0.83×)} & \textbf{0.153 (0.36×)}\\
\hline
\end{tabular}
\label{TAB:RMSEOF}
\end{table*}

\begin{figure}[t!]
    \centering
    \subfloat[X-axis translation]
    {
        \includegraphics[width=1\linewidth]{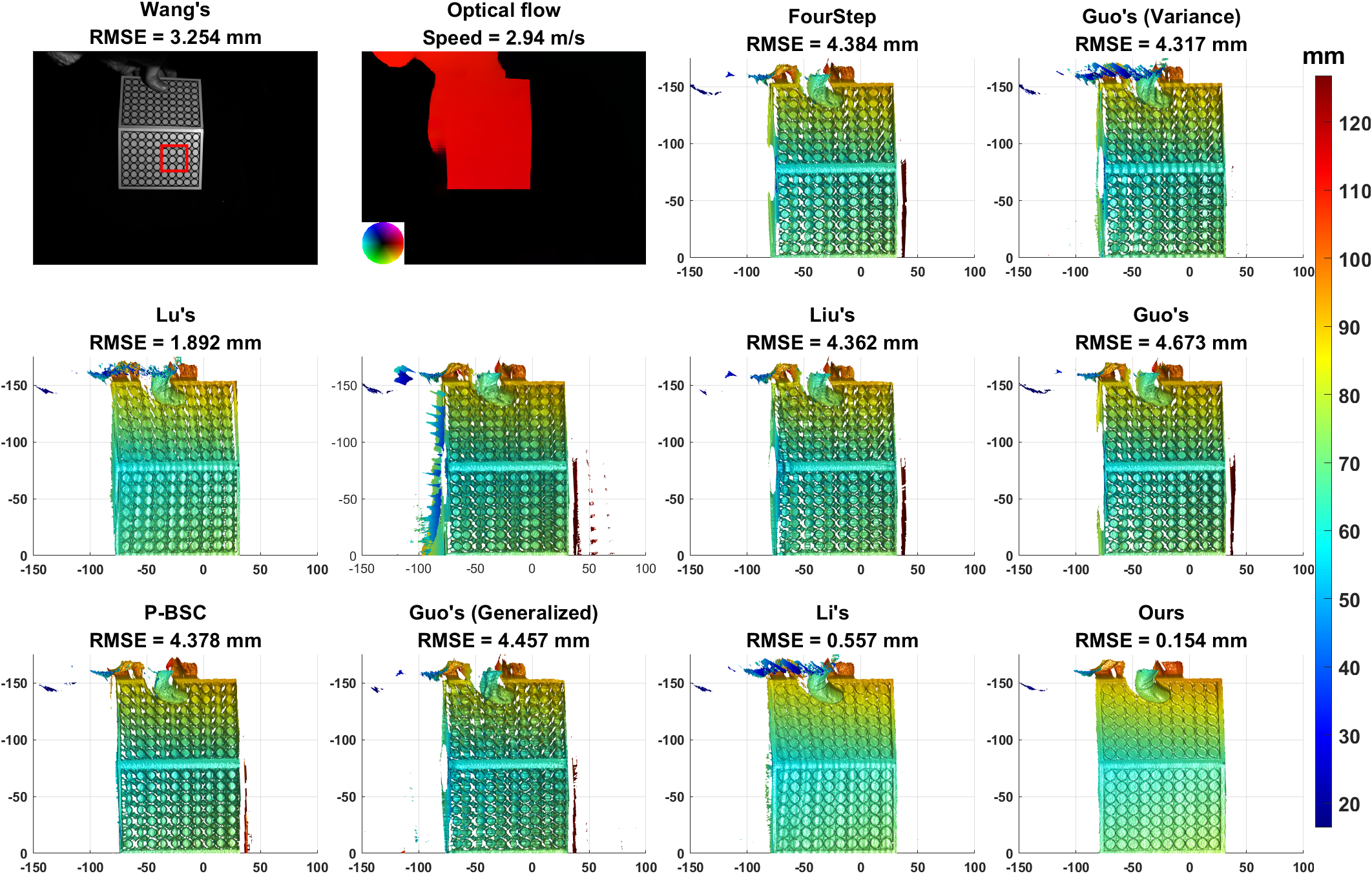}
    }
    \\
    \subfloat[Z-axis translation]
    {
        \includegraphics[width=1\linewidth]{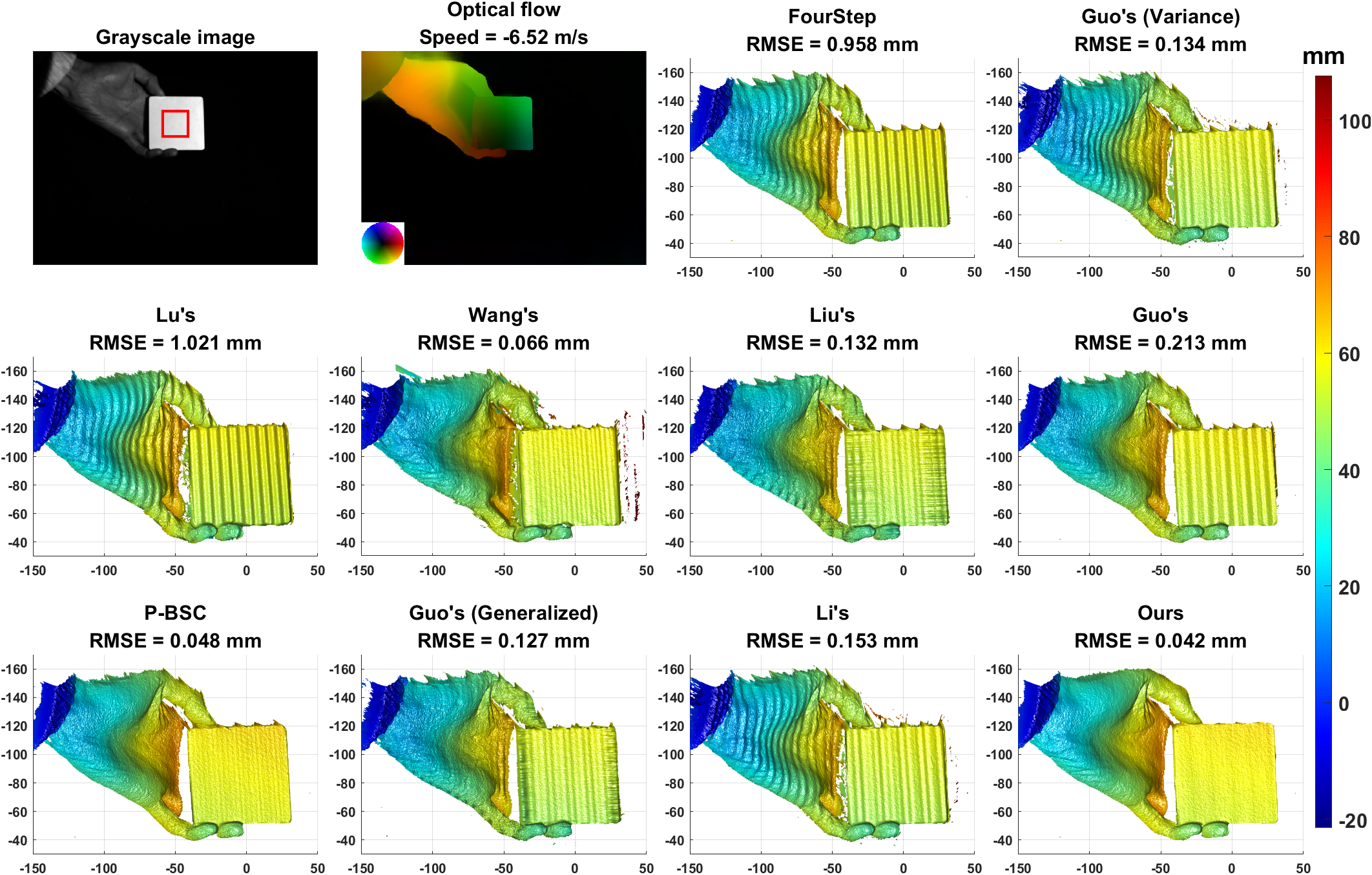}
    }
    \\
    \subfloat[Z-axis rotation]
    {
        \includegraphics[width=1\linewidth]{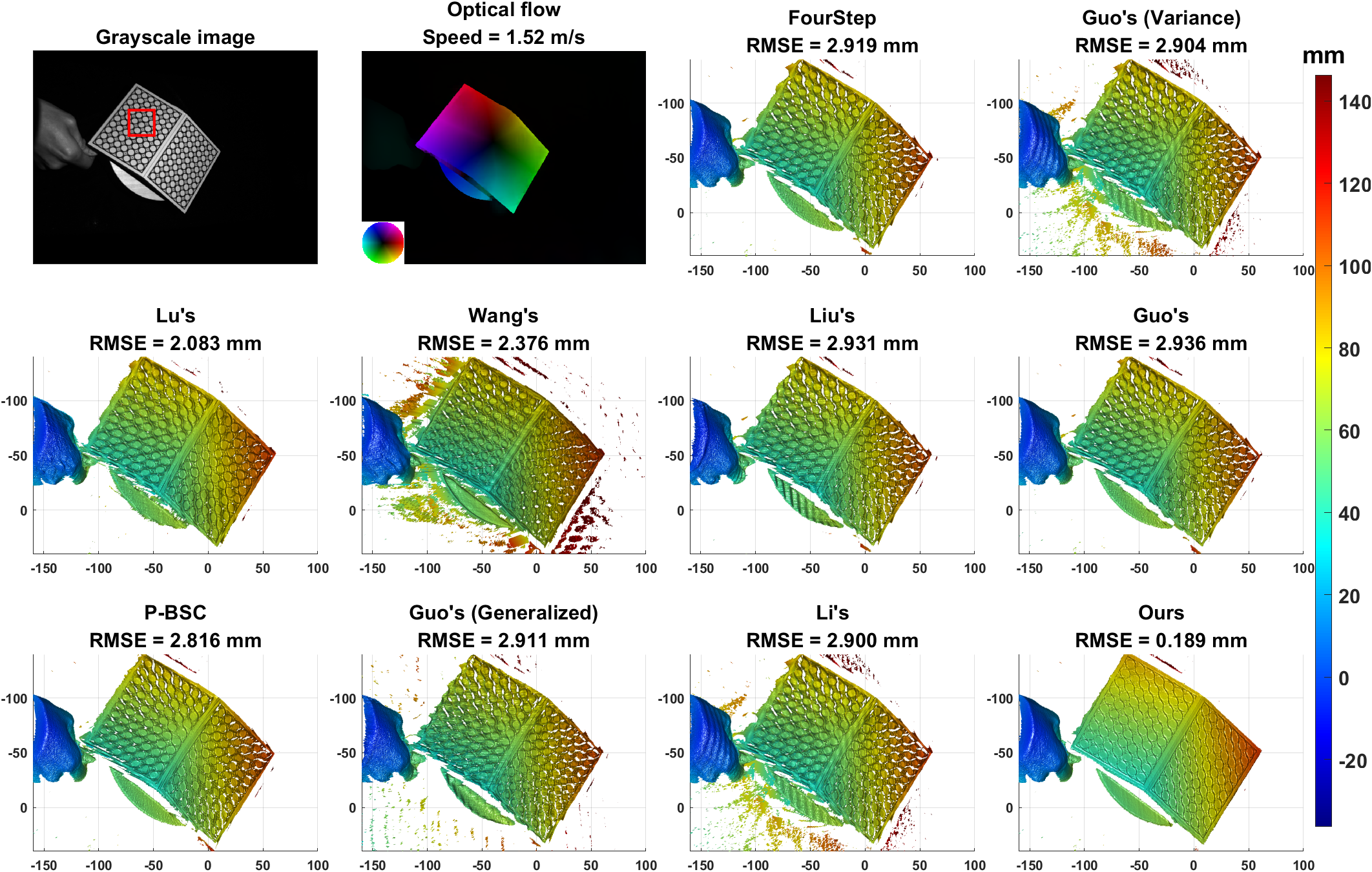}
    }
    \caption{Reconstructed point clouds under a) X-axis translation, b) Z-axis translation, and c) Z-axis rotation. This figure visualizes a frame slice of the measurement results in Fig.~\ref{FIG:FigSpeedVSRMSE} and Table~\ref{TAB:RMSE}. We compute the RMSE within the red rectangular area. As the results show, existing methods exhibit significant residual ghosting artifacts under X-axis translation and Z-axis rotation, as well as obvious ripple-like distortions under Z-axis translation. In contrast, our method effectively suppresses both errors to visually imperceptible levels. Please note that due to the presence of multi-path errors, the RMSE when using the standard plate with circular markers is obviously higher than that of the white standard plate. However, multi-path-induced errors fall beyond the scope of this study.} 
    \label{FIG:FigOneFrame}
\end{figure}
We meticulously conducted a comprehensive quantitative evaluation. We compared the performance of our method and competing approaches in handling 6-DoF motion scenarios involving translation along the X-, Y-, and Z-axes and rotation around the X-, Y-, and Z-axes. In the Z-axis translation experiment, we specifically focused on ripple-like distortions; therefore, the imaging target was a uniform white standard flat plate to eliminate the influence of surface texture. For other motion scenarios, our primary concern was motion artifacts; hence, we employed a calibration board with an array of circular markers to make ghosting artifacts clearly observable. It should be noted that complex textures can also introduce errors due to multi-path effects, which explains the presence of substantial errors (0.15 mm) even when the object remained stationary during translation along X- and Y-axes and rotation around X-, Y-, and Z-axes experiments. However, multi-path-induced errors fall beyond the scope of this study. The frame rate for the camera capturing is set to 454 fps. We measured the moving target at a distance of 500~mm by using our RPSP-AM~(8 frames and 2 uniform patterns) and some representative methods for comparison: 1) traditional four-step PSP~(4 frames), 2) Carré's~\cite{carre1966installation}, 3) Guo's~\cite{guo2013phase} phase shift estimation from variances~(4 frmaes), 4) Lu's~\cite{lu2013new} method (4 frames), 5) Wang's~\cite{wang2018motion} Hilbert transform compensation~(4 frames), 6) Liu's~\cite{liu2019real} method (8 frames), 7) Guo's~\cite{guo2021real} phase frame sum method~(5 frmaes), 8) our previous proposed P-BSC~\cite{zhang2024binomial} (8 frames), 9) Guo's~\cite{guo2025generalized} generalized phase shift deviation estimation (6 frames), 10) Li's~\cite{li2025mrpca} motion-resistant PSP~(4 frmaes). In Lu’s original scheme~\cite{lu2013new}, three markers are placed on the object, which may be difficult to achieve in practical measurements. Therefore, we reproduced their method for comparison by estimating the rotation matrix and translation vector using the optical flow map. Our measurement lasted for 1400 frames (about 3 seconds) for each scenario. We selected a $70\times70$ rectangular window in each depth map and fit the plane with the random sample consensus algorithm as the ground truth, then we computed the root mean squared error (RMSE) within the selected regions. 

The trend of RMSE changing with motion speed in each scenario is depicted in Fig.~\ref{FIG:FigSpeedVSRMSE}. Our method maintained superior accuracy across all tested motion scenarios and speed ranges, consistently achieving the lowest RMSE without significant influence from increasing motion speed. Under translation along X- and Y-axes and rotation around X- and Y-axes, Li's~\cite{li2025mrpca} method demonstrated the second-lowest RMSE with stable accuracy across the speed range. However, its RMSE increased markedly with speed under Z-axis translation and rotation. P-BSC~\cite{zhang2024binomial} exhibited stable accuracy under Z-axis translation across the experimental speed range, where it performed the closest to our method. In contrast, all other comparison methods showed a significant rise in RMSE as motion speed increased, across all motion scenarios.

The reconstructed point clouds under representative motion scenarios (X- and Z-axis translation, Z-axis rotation) are visualized in Fig.~\ref{FIG:FigOneFrame}. Under Z-axis translation, which primarily induces ripple-like distortions, the point clouds generated by our method and P-BSC~\cite{zhang2024binomial} remain notably flat, whereas all other methods exhibit obvious residual ripples. In contrast, under X-axis translation and Z-axis rotation where ghosting artifacts dominate, our method successfully reconstructs a high-fidelity point cloud of the calibration board, while the results of other comparison methods are compromised by significant ghosting artifacts.

We computed the average RMSE across different motion speeds and list in Table~\ref{TAB:RMSE}. Quantitative assessments indicate that under arbitrary 6-DoF motion conditions, our RPSP-AM exhibits superior performance, while existing methods are typically only effective for specific types of motion. Among all the methods, our RPSP-AM achieved the lowest average RMSE of 0.153~mm, reducing the motion errors to 0.09$\times$ compared with four-step phase-shifting method. Li's~\cite{li2025mrpca} demonstrated the second-lowest average RMSE of 0.367~mm (0.21~$\times$).

\textbf{Line-of-sight motion} (ripple-like distortion): under the condition of translation along the Z axis, ripple-like distortions become dominant. Our method delivered the lowest RMSE of 0.046~mm, suppressing the ripple-like distortions to 0.08$\times$ compared with four-step phase-shifting method. Lu's~\cite{lu2013new} method is tailored for FPM, so it was unable to effectively suppress the ripple-like distortions. The RMSE reached 0.538~mm, which shows no significant difference from the 0.551~mm of the four-step phase-shifting method. P-BSC~\cite{zhang2024binomial} achieved the second-lowest performance, with an RMSE of 0.054~mm (0.09$\times$), demonstrating accuracy that was very close to ours. The other approaches effectively suppressed ripple-like distortions, ranging from 0.213~mm (0.39$\times$) to 0.089~mm (0.16$\times$). 

\textbf{Focal plane motion} (ghosting artifacts): under translation along the X- and Y-axes and rotation around the X-, Y-, and Z-axes, ghosting artifacts dominate due to image misalignment. Still, our method delivered the lowest RMSE, recorded at 0.171~mm, 0.142~mm, 0.163~mm, 0.171~mm, and 0.222~mm, respectively. This corresponds to a remarkable reduction in error to 0.06$\times$, 0.08$\times$, 0.09$\times$, 0.07$\times$, and 0.27$\times$ compared to the conventional four-step phase-shifting baseline. Li's~\cite{li2025mrpca} method achieved the second-lowest performance in scenarios involving translation along X- and Y-axes and rotation around X- and Y-axes, achieving RMSEs of 0.381 mm (0.13×), 0.224 mm (0.12×), 0.268 mm (0.14×), and 0.375 mm (0.15×), respectively. However, it fails to do so under rotation around Z-axis. This is because Li's~\cite{li2025mrpca} method relies on the assumption of a globally consistent translation vector, which is severely violated in the presence of rotation around Z-axis. Lu's~\cite{lu2013new} method achieved RMSE of 0.454~mm under rotation around Z-axis, suppressing the ghosting artifacts to 0.56$\times$. This is because Lu's~\cite{lu2013new} method assumes a global consistent rotation matrix, which enables an effective description of the rotation around the Z-axis.

To facilitate an intuitive performance comparison across methods, the data from Table~\ref{TAB:RMSE} is summarized as a spider plot in Fig.~\ref{FIG:FigSpeedVSRMSESpider}. As observed, our method achieves the lowest RMSE across all 6-DoF motions, reducing the RMSE by more than 10× compared to the conventional four-step phase-shifting method. Existing approaches, by contrast, are typically effective only for specific motion types. For instance, methods such as Carré's~\cite{carre1966installation}, Guo's (Variance)~\cite{guo2013phase}, Wang's~\cite{wang2018motion}, Liu's~\cite{liu2019real}, Guo's~\cite{guo2021real}, P-BSC~\cite{zhang2024binomial}, and Guo's (Generalized)~\cite{guo2025generalized} are designed to address motion-induced phase deviations and thus perform well mainly for Z-axis translation. In contrast, Lu's~\cite{lu2013new} method compensates only for in-plane translation and rotation, making it ineffective in handling Z-axis translation. This indicates our method's comprehensive capability to thoroughly and significantly suppress motion errors, including both ripple-like distortions and ghosting artifacts, across arbitrary motion scenarios. 

Further, in order to more accurately determine the contribution of I-BSC to suppressing ripple-like distortions, we performed optical-flow-based image alignment for all the comparison methods, thereby eliminating ghosting artifacts and leaving only the ripple-like distortions. RAFT~\cite{teed2020raft} is adopted to align the fringe image sequence for enhancing the comparison methods. The comparison results between our method and the optical flow enhanced version of existing techniques are listed in~\ref{TAB:RMSEOF}. After optical-flow-based image alignment, all methods demonstrate improved accuracy, indicating that the image alignment operation significantly reduces ghosting artifacts. Specifically, the average RMSEs of four-step PSP, Carré's~\cite{carre1966installation}, Guo's~\cite{guo2013phase} (variance), Lu's~\cite{lu2013new}, Wang's~\cite{wang2018motion}, Liu's~\cite{liu2019real}, Guo's~\cite{guo2021real}, P-BSC~\cite{zhang2024binomial}, Guo's~\cite{guo2025generalized} (generalized), and Li's~\cite{li2025mrpca} methods are reduced from 1.742~mm, 2.031~mm, 1.665~mm, 0.621~mm, 1.380~mm, 1.688~mm, 1.590~mm, 1.580~mm, 1.744~mm, 0.367~mm, to 0.429~mm, 0.267~mm, 0.183~mm, 0.253~mm, 0.881~mm, 0.187~mm, 0.186~mm, 0.159~mm, 0.184~mm, 0.183~mm, respectively. Still, our method maintains superior comprehensive performance under 6-DoF motion, proving that our I-BSC is of state-of-the-art performance in suppressing ripple-like distortions. Among all the methods, P-BSC~\cite{zhang2024binomial} achieved the second-lowest average RMSE of 0.163~mm, demonstrating performance that was very close to ours. Lu's~\cite{lu2013new} method showed the second-lowest RMSE of 0.230~mm under rotation around Z-axis. The supplementary document provides visualizations of the optical-flow-enhanced results of the comparison methods, including the spider plot, error-speed plot, and point cloud.
\begin{figure*}[t!]
    \centering
    \includegraphics[width=0.85\linewidth]{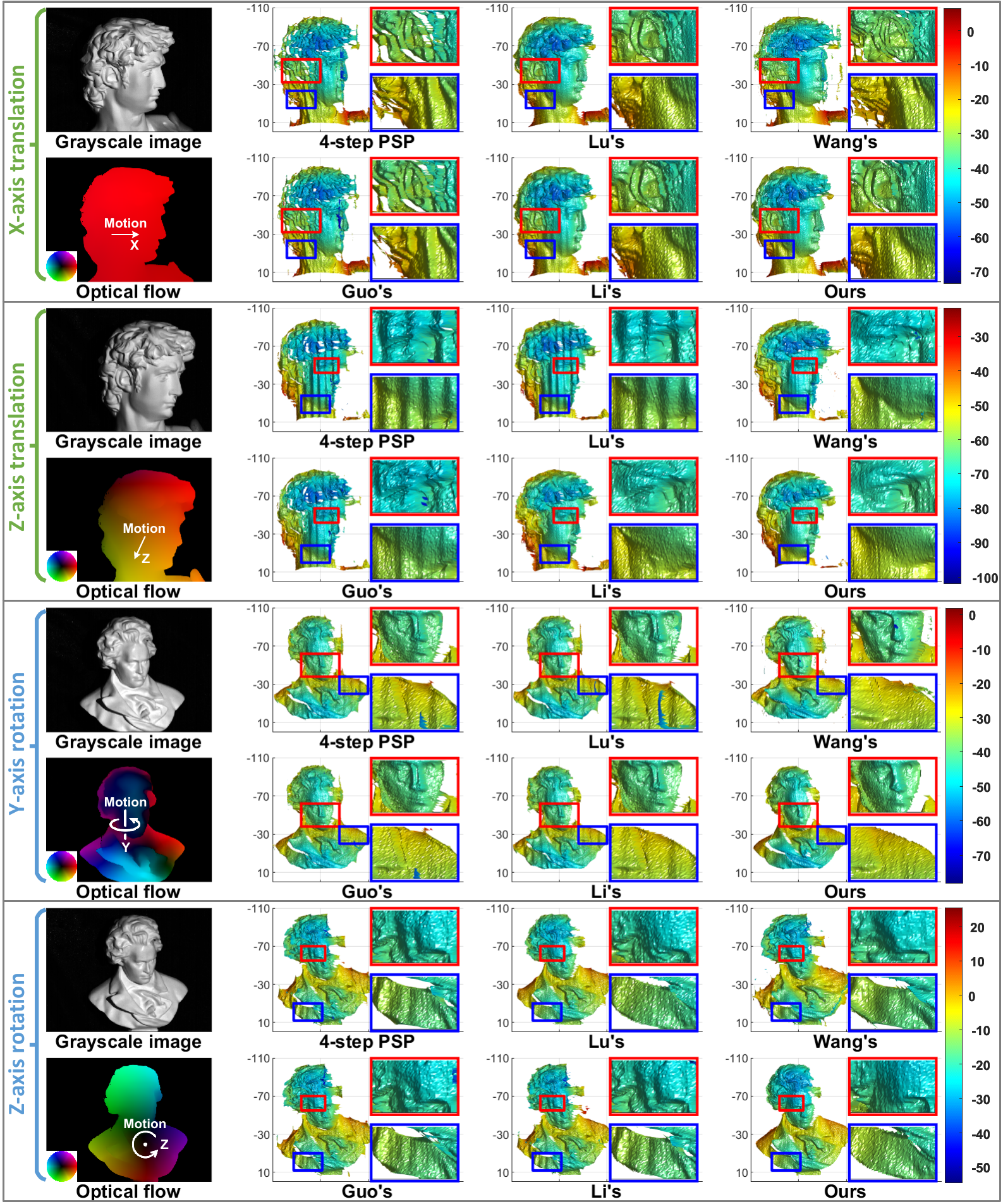}
    \caption{Reconstructed results (unit: mm) of measured objects under different rigid motion scenarios: row 1-2 (X-axis translation): Wang's~\cite{wang2018motion} and Guo's~\cite{guo2021real} methods exhibit notable ghosting artifacts; row 3-4 (Z-axis translation): Lu's~\cite{lu2013new}, Wang's~\cite{wang2018motion}, and Guo's~\cite{guo2021real} methods demonstrate ripple-like distortions; row 5-6 (Y-axis rotation): Lu's~\cite{lu2013new}, Wang's~\cite{wang2018motion}, and Li's~\cite{li2025mrpca} methods exhibit both ghosting and ripples, while Guo's~\cite{guo2021real} shows only ghosting; row 7-8 (Z-axis rotation): Wang's~\cite{wang2018motion}, Guo's~\cite{guo2021real}, and Li's~\cite{li2025mrpca} methods showcase ghosting artifacts. In contrast, our RPSP-AM reconstructs high-fidelity motion-error-free 3D surfaces across all the above conditions.}
    \label{FIG:FigRigidMotion}
\end{figure*}

\subsection{Qualitative Evaluation on Complex Scenarios}
\label{sectionB}

\begin{figure*}[t!]
    \centering
    \includegraphics[width=0.85\linewidth]{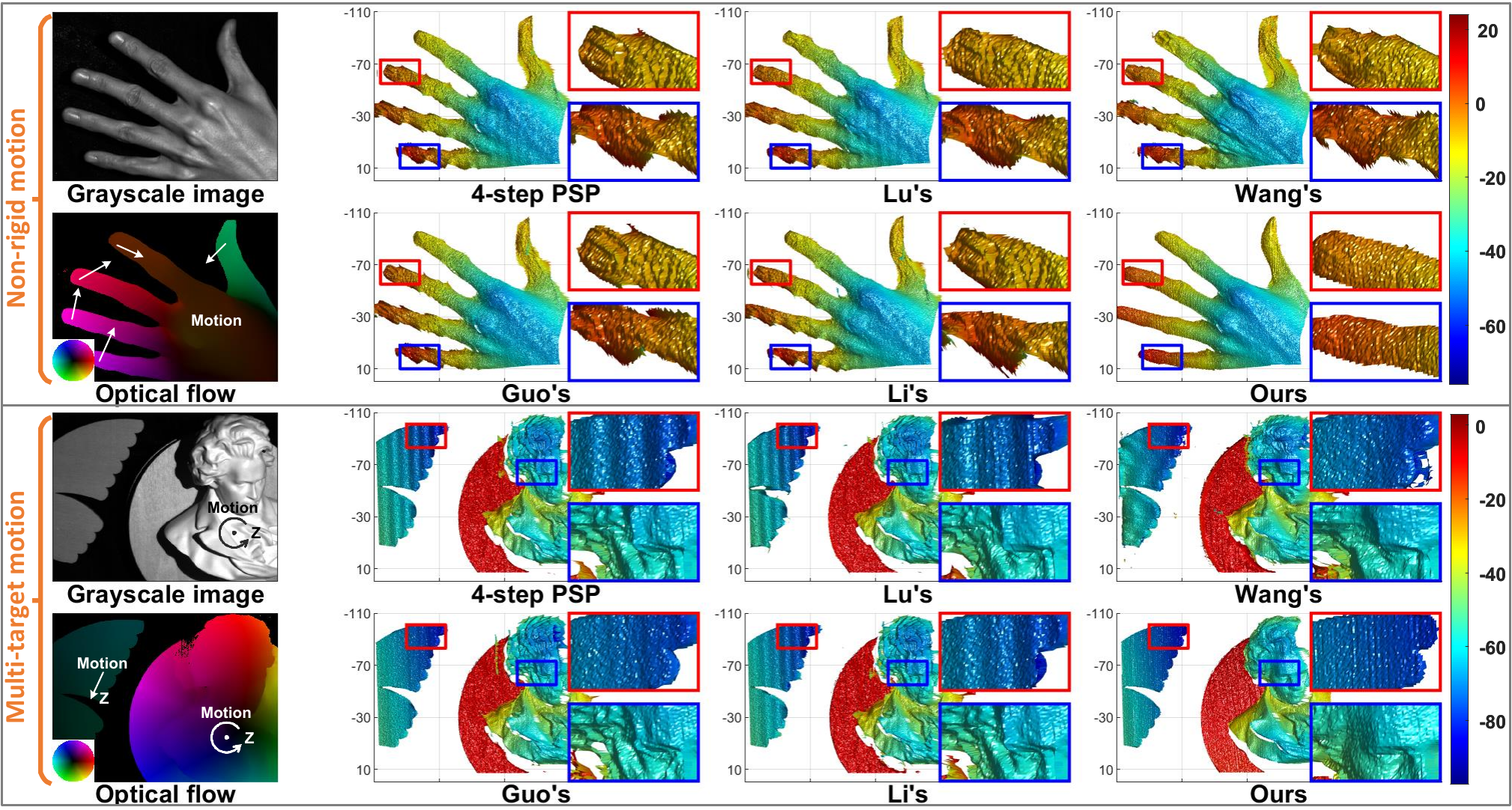}
    \caption{Reconstructed results (unit: mm) of measured objects under non-rigid motion and multi-target motion: row 1-2: non-rigid motion; row 3-4: multi-target motion.  Lu's~\cite{lu2013new}, Wang's~\cite{wang2018motion}, Guo's~\cite{guo2021real}, and Li's~\cite{li2025mrpca} methods exhibit ripple-like distortions and/or ghosting artifacts, whereas our RPSP-AM reconstructs high-fidelity motion-error-free point clouds.}
    \label{FIG:FigComplexMotion}
\end{figure*}

To demonstrate the generality of BSC, we employed our RPSP-AM and five comparitive methods 1) traditional four-step PSP, 2)  Lu's~\cite{lu2013new} method, 3) Wang's~\cite{wang2018motion} Hilbert transform compensation, 4) Guo's~\cite{guo2021real} phase frame sum method, and 5) Li's~\cite{li2025mrpca} motion-resistant PSP to scan objects performing rigid motion (such as translation along the X- and Z-axes, and rotation around the Y- and Z-axes). The frame rate for the camera capturing is set to 90 fps. The measurement distance is about 600~mm and the field of view is 150~mm$\times$120~mm. The results are shown in Fig.~\ref{FIG:FigRigidMotion}. The performance of the comparison methods varied significantly across different motion types. Under motions inducing ghosting artifacts (X-axis translation),  Lu's~\cite{lu2013new} and Li's~\cite{li2025mrpca} methods effectively suppress errors, in contrast to Wang's~\cite{wang2018motion} and Guo's~\cite{guo2021real} methods, which exhibit noticeable artifacts. Conversely, under Z-axis translation dominated by ripple-like distortions,  Wang's~\cite{wang2018motion}, Guo's~\cite{guo2021real}, and Li's~\cite{li2025mrpca} methods perform well, while Lu's~\cite{lu2013new} method shows significant ripple-like distortions. Under Y-axis rotation, Lu's~\cite{lu2013new}, Wang's~\cite{wang2018motion}, Guo's~\cite{guo2021real}, and Li's~\cite{li2025mrpca} methods all show slight artifacts. Under Z-axis rotation, Lu's~\cite{lu2013new} method is effective in suppressing ghosting artifacts, whereas the methods of Wang~\cite{wang2018motion}, Guo~\cite{guo2021real}, and Li~\cite{li2025mrpca} all showcase significant residual artifacts. Moreover, it can be seen that, by conducting Hilbert transform on the captured images, Wang's~\cite{wang2018motion} method is effective in mitigating motion error, yet inevitably causing spectrum truncation, thereby inducing Gibbs phenomenon at depth variation areas. Therefore, we can conclude that our BSC with pixel-wise property is more robust compared with non-pixel-wise methods when dealing with complex surface or depth discontinuous scenes, showcasing strong resolving ability in detail. 

We also tested objects undergoing complex motions (such as non-rigid and multi-target motion), with the results displayed in Fig.\ref{FIG:FigComplexMotion}. Residual ripple-like distortions or ghosting artifacts exist in the results of Lu's~\cite{lu2013new}, Wang's~\cite{wang2018motion}, and Guo's~\cite{guo2021real} methods, as they solely compensate for either ghosting artifacts or ripple-like distortions. Li’s~\cite{li2025mrpca} method addresses both ghosting artifacts and ripple-like distortion but is limited to globally consistent motion, as it models in-plane movement using a global $2\times1$ translation vector and assumes a global consistent phase shift for motion along the Z-axis, leading to errors under rotational, non-rigid, or multi-object motions. Our RPSP-AM comprehensively eliminates both ghosting artifacts and ripple-like distortions, achieving robust 3D reconstruction for arbitrary motion, including translation and rotation in any direction, non-rigid motion, and multi-target motion. 

In addition, the optical flow estimation can be integrated into existing motion error compensation techniques to eliminate the ghosting artifacts. Thus, RAFT is adopted to align the fringe image sequence for enhancing the comparison methods, significantly suppressing the ghosting artifacts. The comparison results between our method and the optical-flow-enhanced version of existing techniques can be found in the supplementary document. Across all the motion conditions, our RPSP-AM consistently demonstrates the most robust performance against ripple-like distortions.

We attached videos in the supplementary materials as visualizations of the reconstruction results of five dynamic scenes, including: translation along X- and Z-axes, rotation around Z-axis, non-rigid deformation, and multi-target movements. Six methods are employed, including: traditional four-step phase shifting, Lu's~\cite{lu2013new}, Wang's~\cite{wang2018motion}, Guo's~\cite{guo2021real}, Li's~\cite{li2025mrpca} methods, and our RPSP-AM. The videos show that all methods perform consistently with the analysis in Fig.~\ref{FIG:FigRigidMotion} and~\ref{FIG:FigComplexMotion} across different scenarios. The results support that our method outperforms the comparison approaches, achieving high-precision 3D reconstruction for arbitrary motion.

\subsection{Intensity Noise Suppression Ability}
\begin{figure}[t!]
    \centering
    \includegraphics[width=1\linewidth]{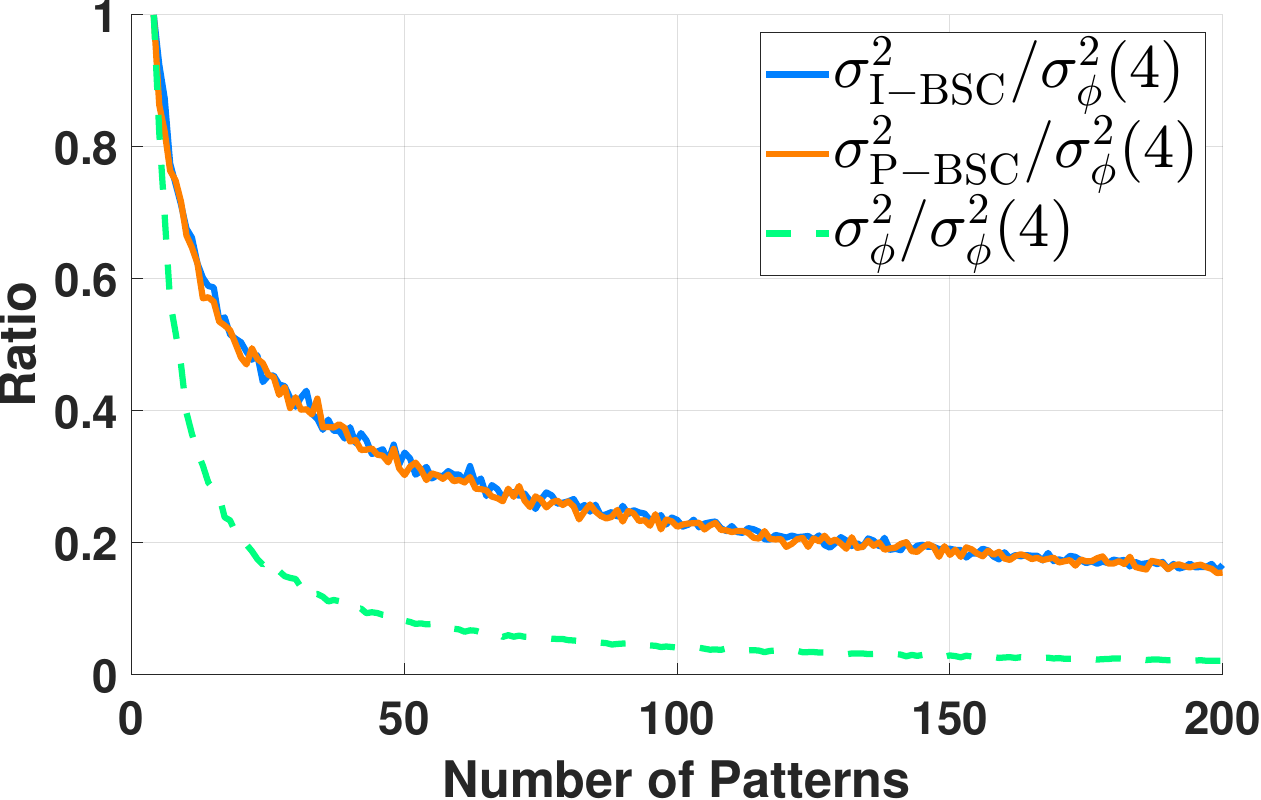}
    \caption{The comparison of intensity noises among P-BSC~\cite{zhang2024binomial}, I-BSC, and traditional $N$-step PSP in simulation.}
    \label{FIG:FigCompareGaussian}
\end{figure}
In a practical FPP system, the intensity of the fringe pattern at each point of the object is acquired by a digital camera and inevitably contaminated by intensity noise~\cite{zuo2018phase}. The intensity noise $\omega_n$ in each captured image originates from various error sources, such as unstable ambient light, projector illumination noise, camera noise, mechanical vibration, etc. According to the central limit theorem, the intensity noise of the captured image can be modeled as an additive Gaussian white noise with zero mean and a variance of $\sigma^2$~\cite{jahne2020release,lv2023modeling}
\begin{equation}\label{EQ:IntensityNoise}
\sigma^2 = \sigma _{dark}^2 + G\left( {I - {I_{dark}}} \right),
\end{equation}
where $I_{dark}$ and $\sigma _{dark}$ represents the dark signal and its variance, $G$ is the camera system gain. Then we can have the variance of Gaussian noise in $N$-step PSP is~\cite{lv2023modeling}
\begin{equation}
\sigma _\phi ^2\left( N \right) = \frac{2}{{N{B^2}}}\left[ {\sigma _{dark}^2 + G\left( {A - {I_{dark}}} \right)} \right]\approx \frac{{2G}}{{NB}}.
\end{equation}
We find that it is difficult to derive a simple form noise model for our P-BSC and I-BSC, thus we conduct a simulation to compare $K$-order P-BSC and I-BSC with $N$-step PSP in terms of intensity noise suppression ability, as shown in Fig.~\ref{FIG:FigCompareGaussian} (we maintain that $K=N-4$ holds true). Consequently, we conduct curve fitting to approximately have a empirical formula as
\begin{equation}
\sigma _{{\rm{I-BSC}}}^2\left( N \right) = \sigma _{{\rm{P-BSC}}}^2\left( N \right) \approx \frac{{G}}{{\sqrt{N}B}}.
\end{equation}
Here, $N$ represents the number of patterns used by P-BSC and I-BSC.

The results show that, P-BSC and I-BSC equivalently mitigate the intensity noise as binomial order $K$ increases, but the noise suppression effects are weaker than traditional $N$-step PSP. Indiscriminately increasing $K$ is uneconomical in reducing intensity noise. Therefore we empirically propose to adopt $K=4$ (8 patterns) for dynamic 3D scanning, which can significantly suppress motion error while also reducing intensity noise.

\begin{figure*}[t!]
    \centering
    \includegraphics[width=0.4\linewidth]{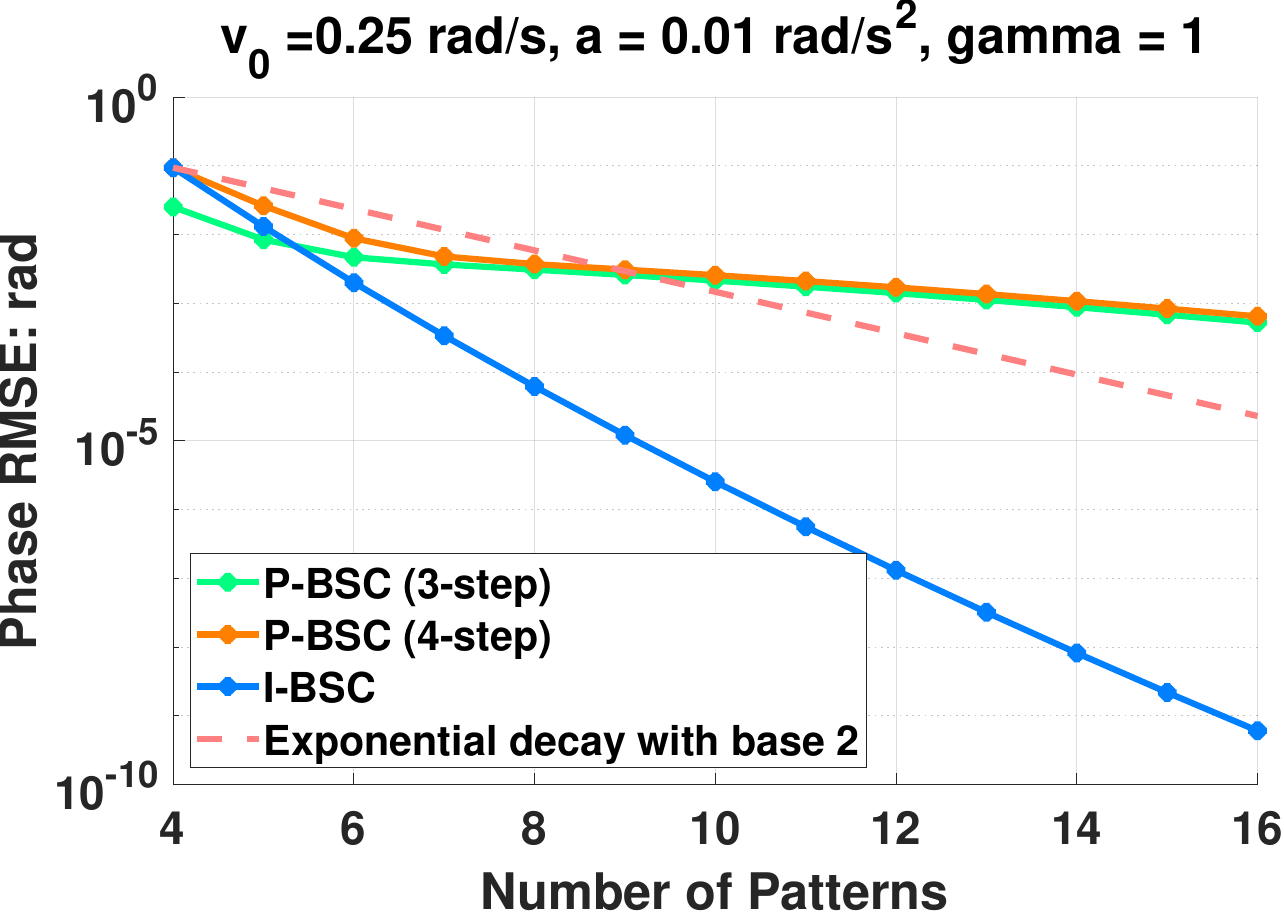}
    \includegraphics[width=0.56\linewidth]{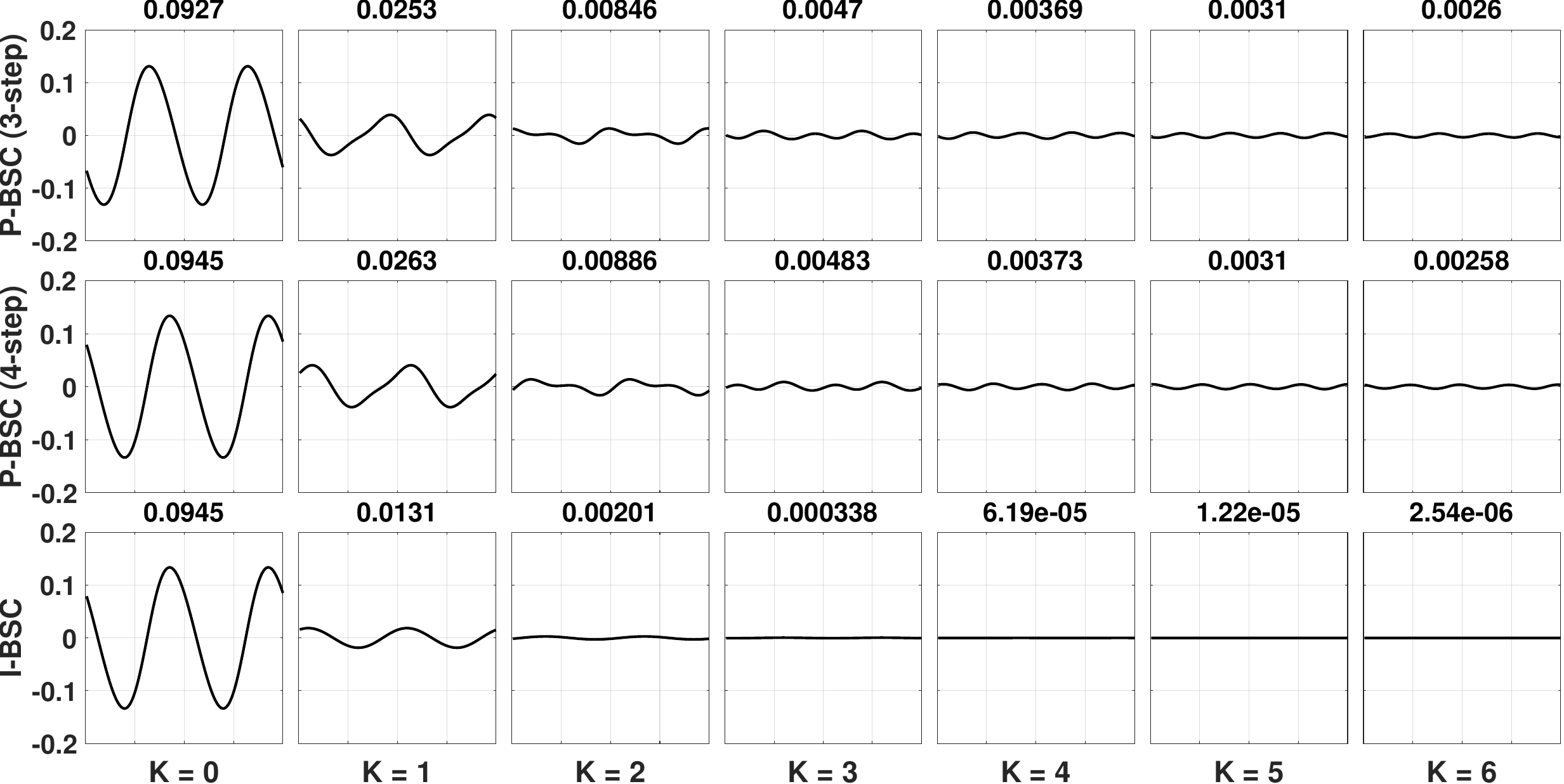}\\
    \vspace{0.2cm}
    \includegraphics[width=0.4\linewidth]{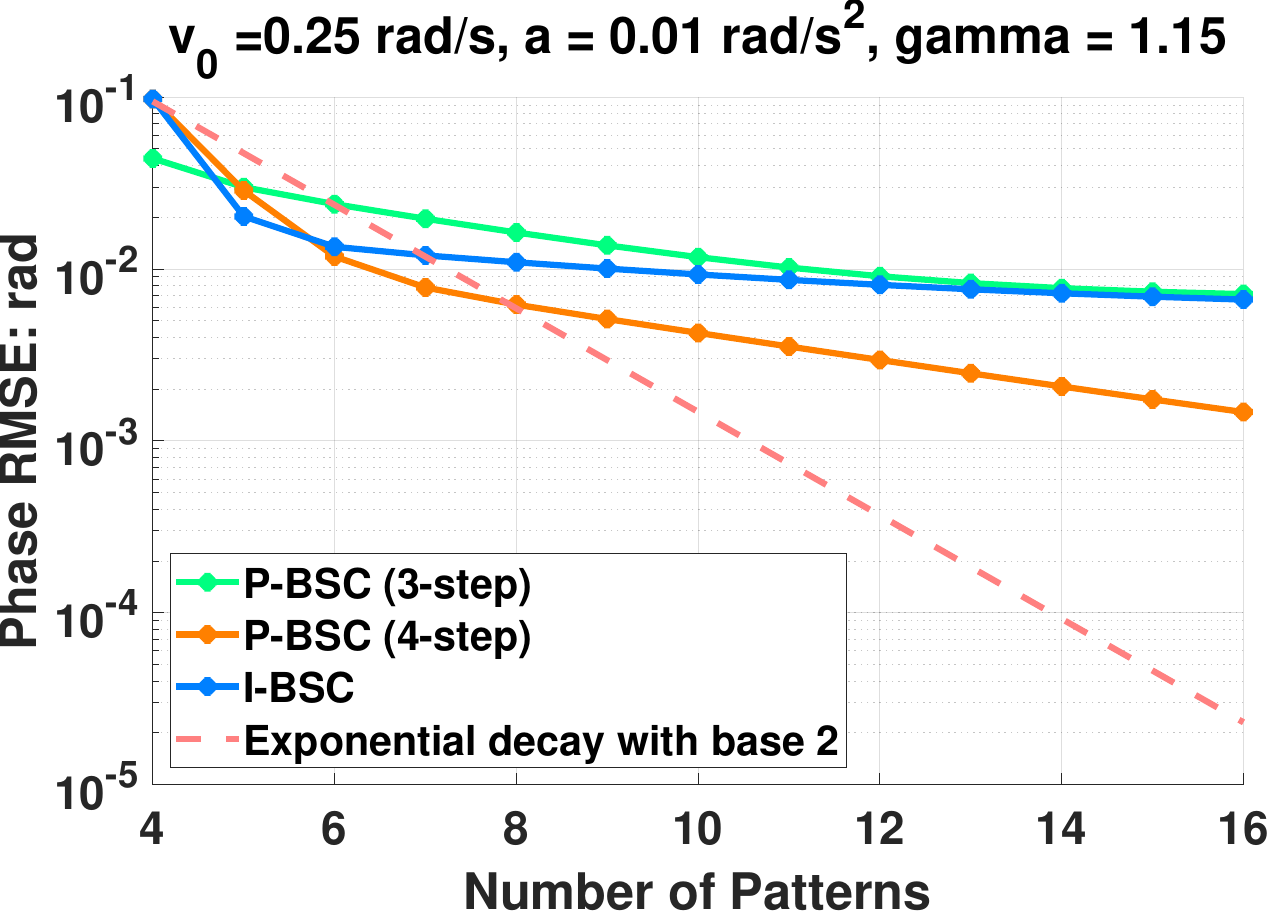}
    \includegraphics[width=0.56\linewidth]{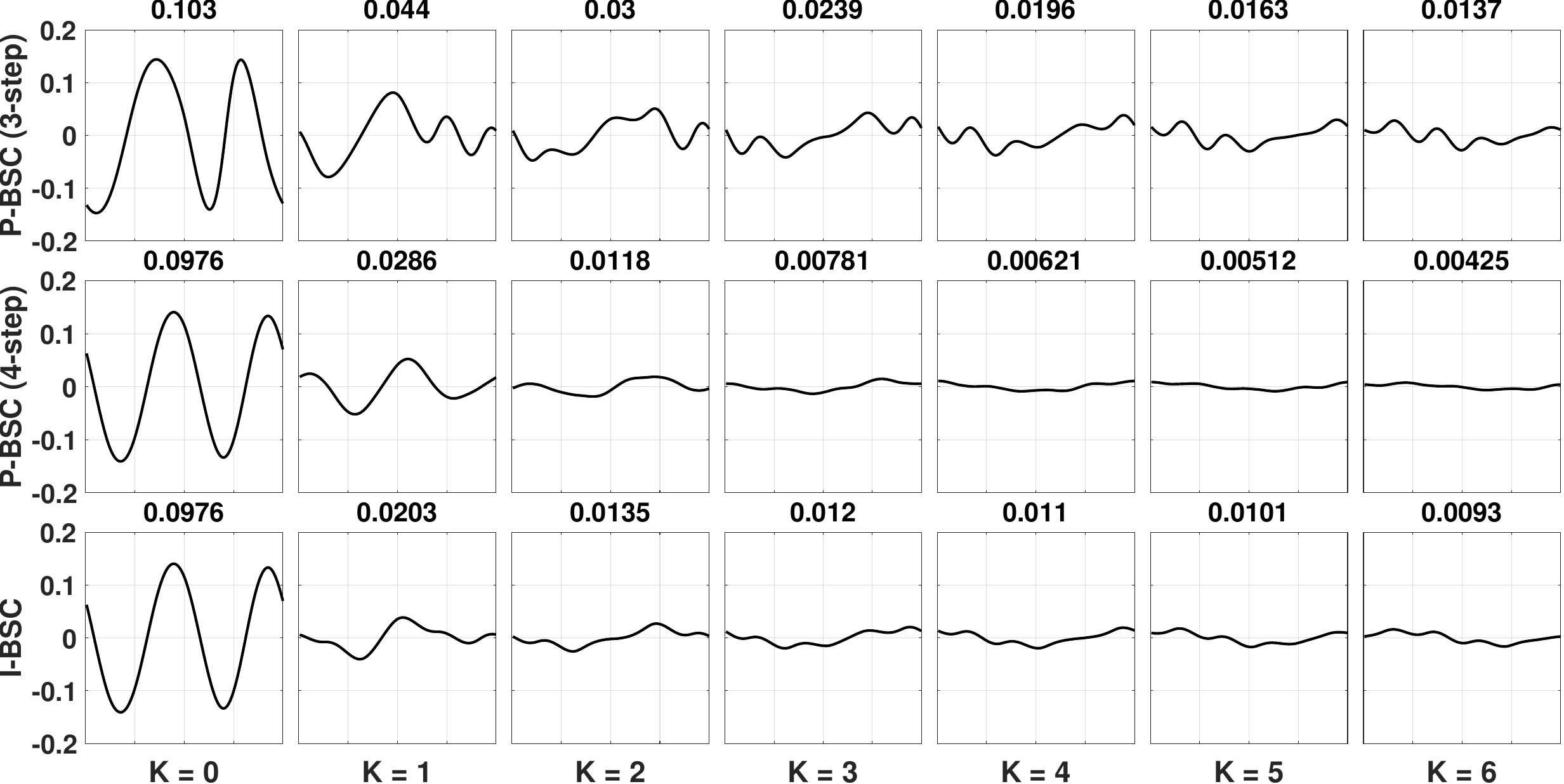}
    \caption{Simulation of motion-induced phase errors in P-BSC~\cite{zhang2024binomial} and I-BSC: Row 1: linear case; Row 2: nonlinear case with gamma = 1.15. The RMSE is labeled above each subfigure.}
    \label{FIG:FigNStepBSC}
\end{figure*}

\begin{figure*}[t!] 
    \centering
    \subfloat[]
    {
        \includegraphics[width=0.45\linewidth]{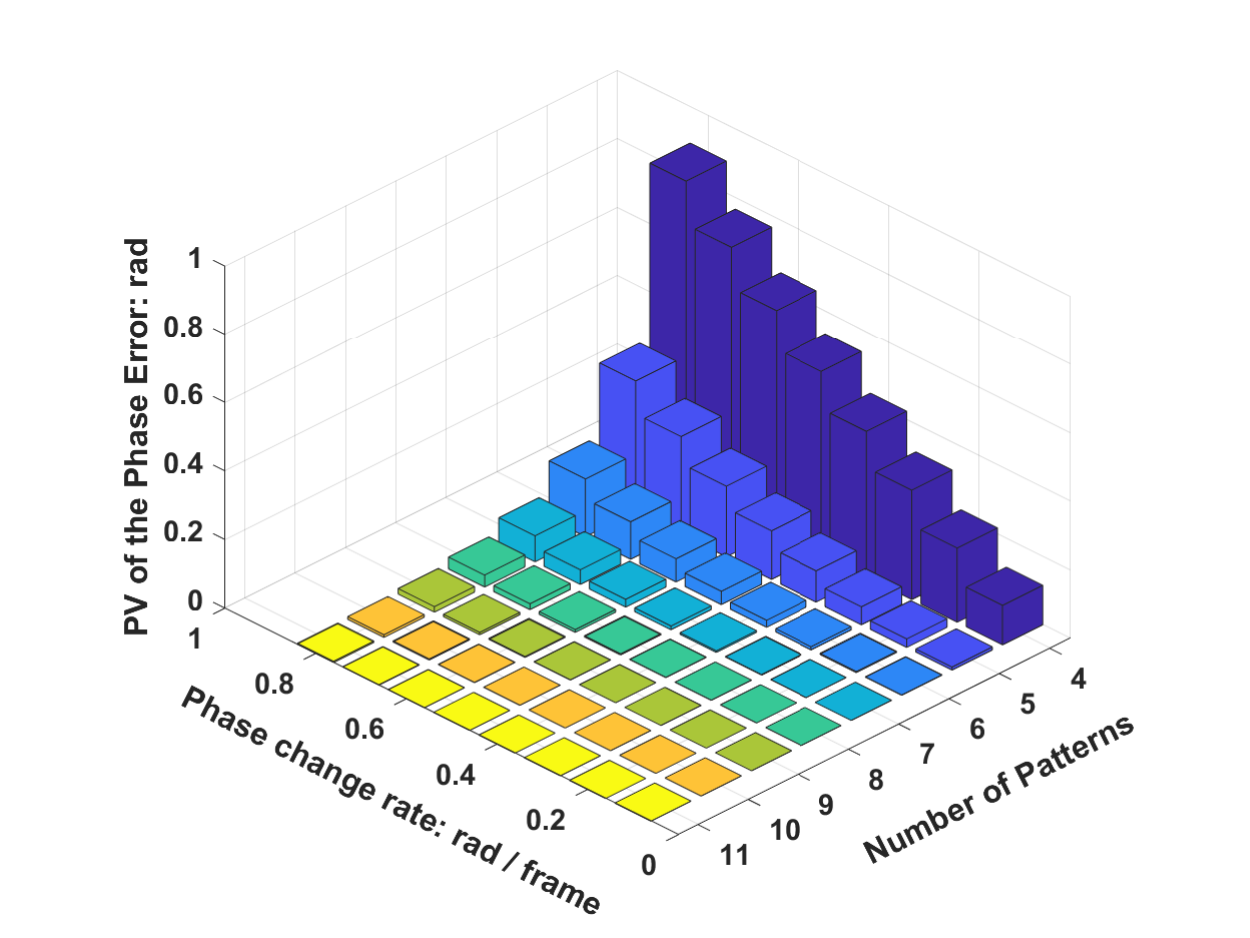}        
    }
    \subfloat[]
    {
        \includegraphics[width=0.5\linewidth]{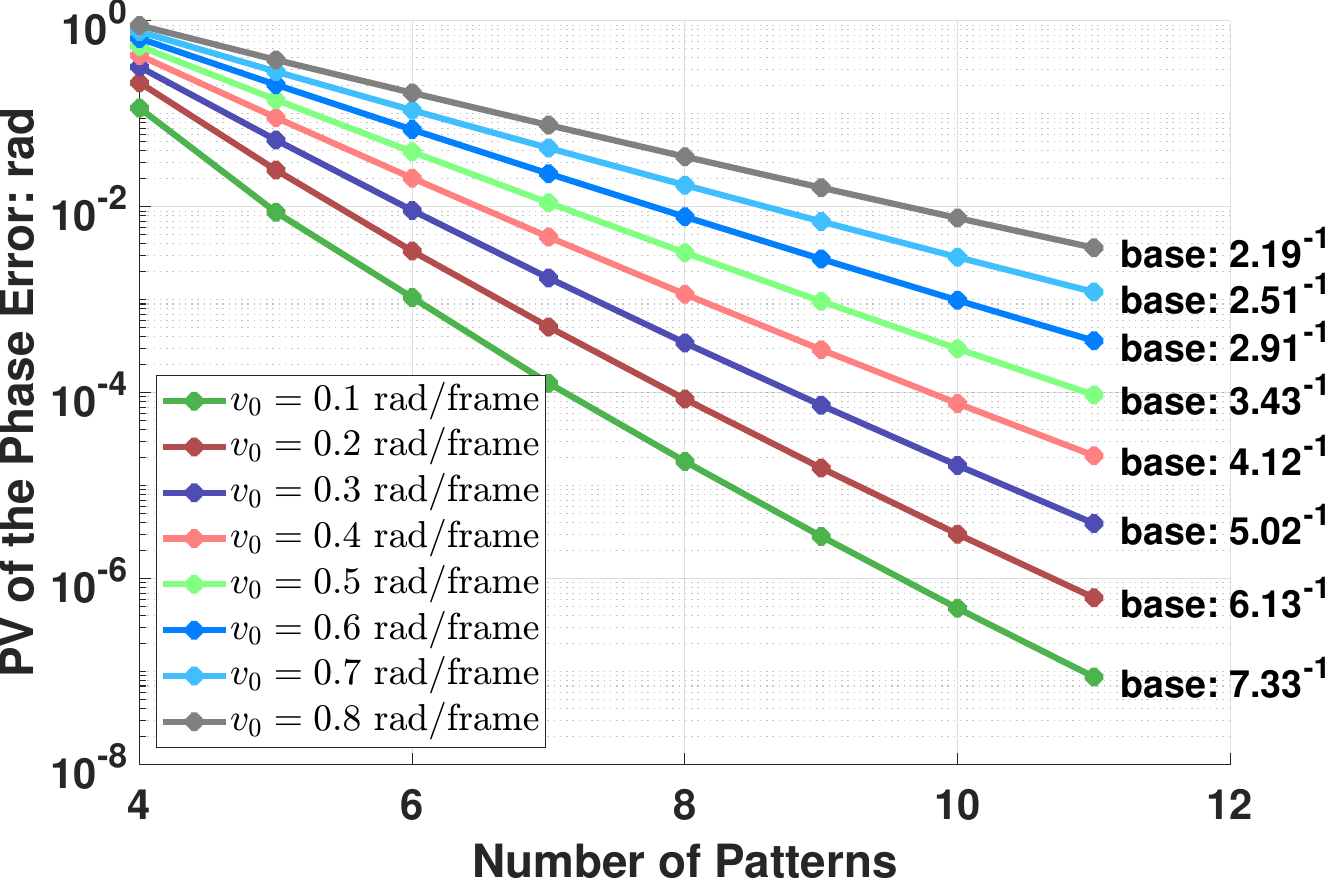}
    }
    \caption{ Simulation results of I-BSC where $v$ ranges from 0.1 to 0.8 rad/frame, the number of patterns ranges from 4 to 11, and $a = 0.01~\mathrm{rad/frame^2}$ remains constant. (a) distribution of PV of phase error with respect to motion-induced phase change rate and the number of patterns used; (b) the trend of PV of phase error with respect to the number of patterns under different phase change rates. The base of the exponential decay obtained from fitting is labeled on the right side of each curve.} 
    \label{FIG:FigErrorVSSpeedVSPatterns}
\end{figure*}

\begin{figure*}[t!] 
    \centering
    \subfloat[]
    {
        \includegraphics[width=0.225\linewidth]{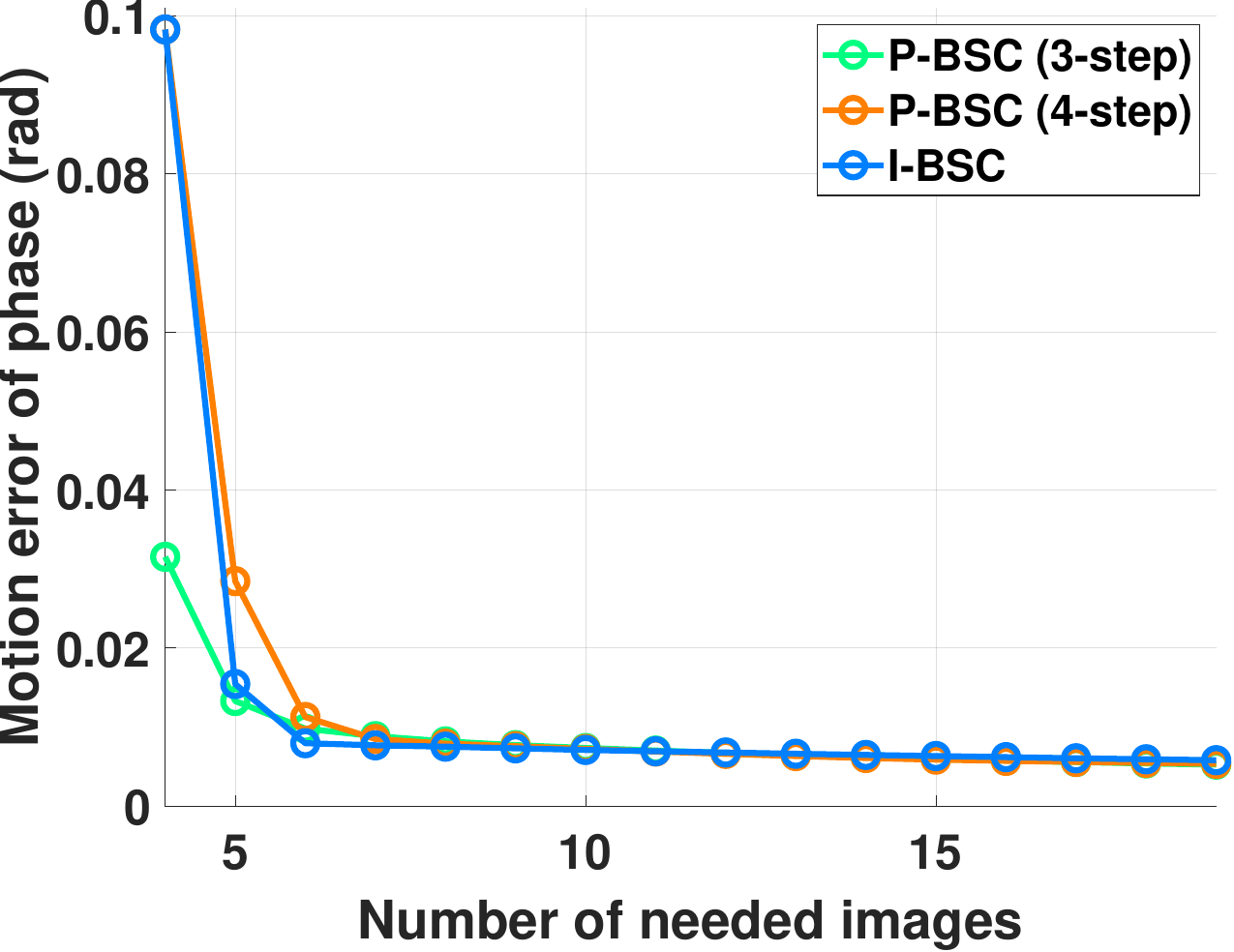}        
    }
    \subfloat[]
    {
        \includegraphics[width=0.24\linewidth]{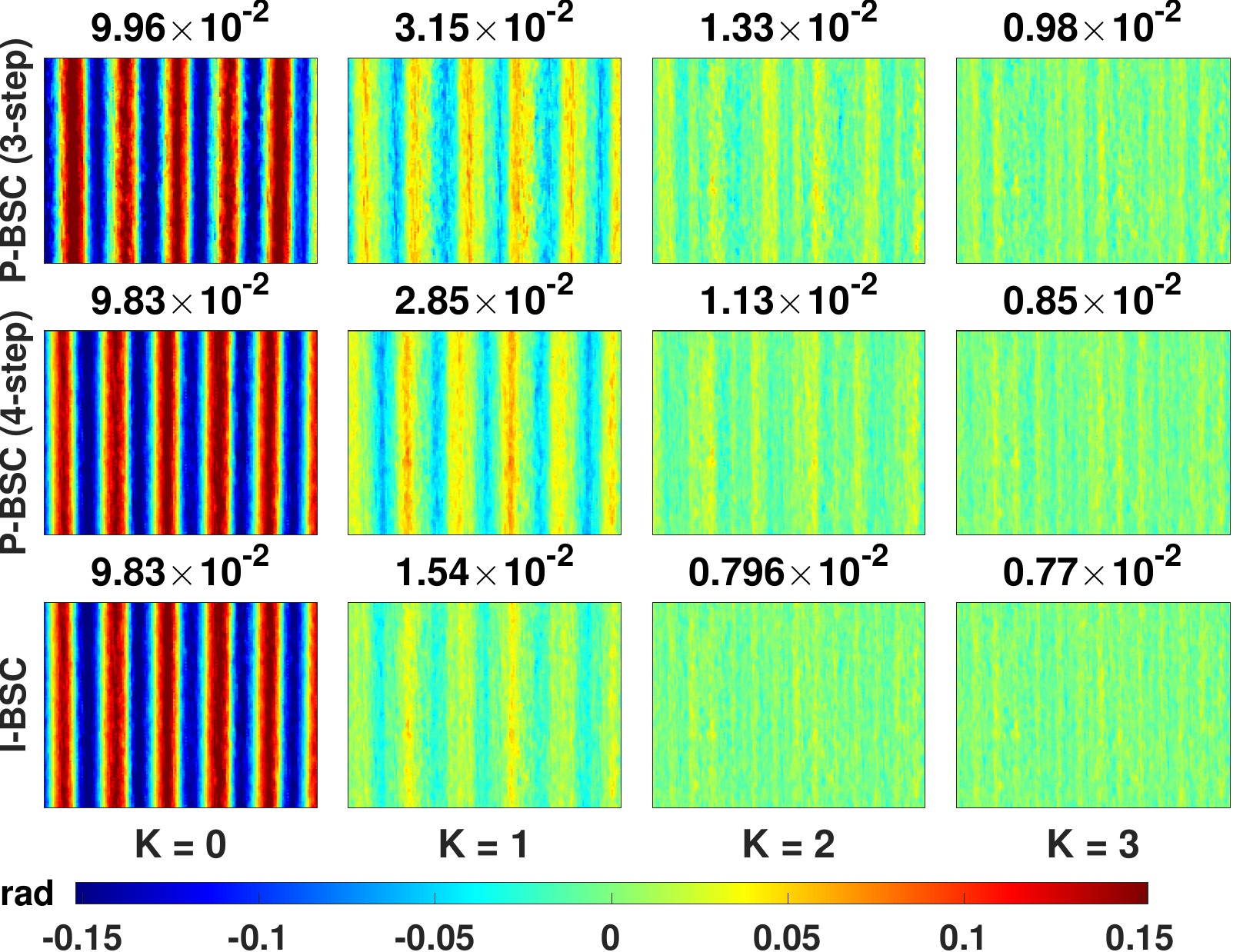}
    }
    \subfloat[]
    {
        \includegraphics[width=0.225\linewidth]{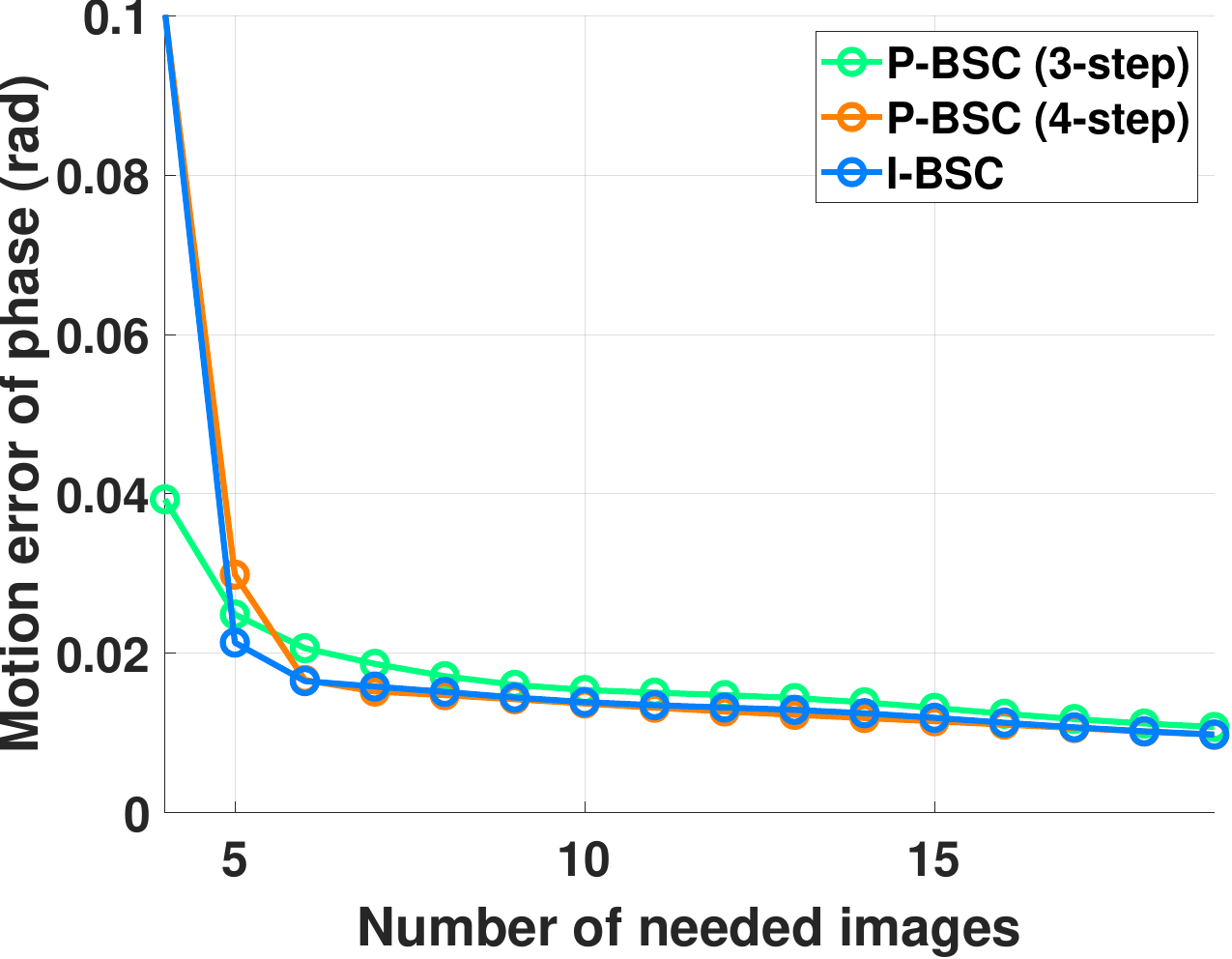}      
    }
    \subfloat[]
    {
        \includegraphics[width=0.24\linewidth]{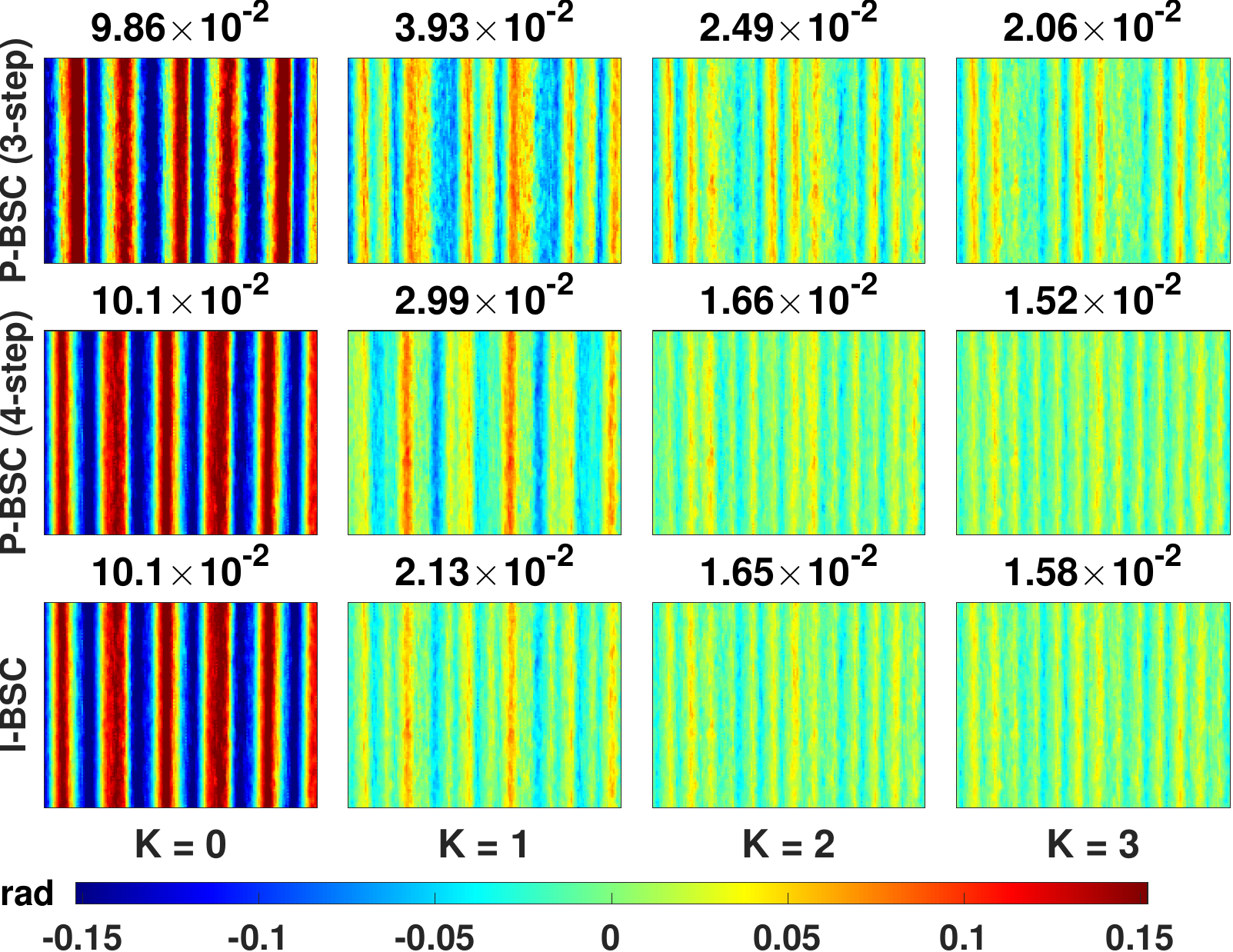}
    }
    \caption{ Real-world experiment on motion-induced phase error of P-BSC~\cite{zhang2024binomial} and I-BSC. The mean absolute error (MAE, labeled above each sub-figure) of phase exponentially diminishes as $K$ increases. (a) and (b) the results with nonlinear rectification; (c) and (d) the results without nonlinear rectification.} 
    \label{FIG:FigPhaseErrorTok}
\end{figure*}

\subsection{Convergence Speed of Motion Error in P-BSC and I-BSC}
\label{sectionConvergenceSpeed}
\subsubsection{Simulations}
We conducted simulations to compare the performance of P-BSC~\cite{zhang2024binomial} and I-BSC. In our simulation, we assumed the dynamic phase shifting error $x_i$ presents a uniform acceleration motion with phase change rate $v_0=0.25~\mathrm{rad/frame}$ and acceleration $a = 0.01~\mathrm{rad/frame^2}$. As shown in row 1 of Fig.~\ref{FIG:FigNStepBSC}, the motion error of I-BSC rapidly converges at a faster rate than exponential decay with base 2. This observation is consistent with the prediction in Eq.~(\ref{EQ:IBSCError2}), which states that due to the combined effects of exponential decay coefficients and higher-order differences, motion errors exhibit a faster convergence rate than exponential decay with base 2. Meanwhile, the motion errors in both the 3-step and 4-step P-BSC also decay quickly but exhibit noticeable residual harmonic errors. This is because the approximation $\tan(\cdot)\approx\cdot$ in Eq.~(\ref{EQ:MotionError4}) is applied $K$ times in the error analysis of P-BSC (Eq.~(\ref{EQ:YangHuiSum4})), resulting in higher-order residual errors when summing. As a result, P-BSC suffers from error accumulation, preventing it from achieving its theoretical accuracy. However, the error accumulation issue does not exist in I-BSC as the approximation is applied only once during the error analysis, and thus leaving no residual errors.

Moreover, we investigated the effect of varying phase-change rates on error suppression performance, as illustrated in Fig.~\ref{FIG:FigErrorVSSpeedVSPatterns}(a). Figure~\ref{FIG:FigErrorVSSpeedVSPatterns}(b) demonstrates that the peak-to-valley (PV) value of motion error decreases almost linearly, with different slopes on the logarithmic scale. A lower phase-change rate leads to a faster reduction in error. It should be noted that when the phase-change rate approaches $\frac{\pi}{2}~\mathrm{rad/frame}$, the inherent phase shifting of the fringe pattern is completely canceled, making our BSC ineffective in this case. Therefore, the proposed BSC is applicable only when the phase-change rate is less than $\frac{\pi}{2}~\mathrm{rad/frame}$.

We evaluated the impact of slight nonlinearity on BSC performance by setting $\gamma = 1.15$, with the results shown in row 2 of Fig.~\ref{FIG:FigNStepBSC}. Specifically, the 3-step P-BSC is significantly affected and is not recommended for this scenario. However, the motion errors in I-BSC and 4-step P-BSC maintain rapid decay when the number of patterns is less than 8, indicating that they retain effective performance in high-speed scenarios. The error suppression capabilities of both P-BSC and I-BSC are impaired in the presence of nonlinearity. Therefore, we recommend using FNR to correct the nonlinearity during the imaging process. The implementation details of FNR can be found in the supplementary document. 

In conclusion, our simulation results indicate that: 1) I-BSC outperforms P-BSC~\cite{zhang2024binomial} in error convergence rate and exponentially reduces motion error; 2) when the phase-change rate is less than $0.8~\mathrm{rad/frame}$, as the number of patterns increases, I-BSC suppresses the motion error at a rate faster than exponential decay with base 2; 3) in the presence of nonlinearity, the performance of both I-BSC and P-BSC~\cite{zhang2024binomial} in mitigating motion errors diminishes.

\subsubsection{Real-world Experiments}
We measured a flat plate translating along Z-axis using our 3-step P-BSC~\cite{zhang2024binomial}, 4-step P-BSC~\cite{zhang2024binomial}, and I-BSC. We conducted experiments separately with and without nonlinear rectification. The results visualized in Fig.~\ref{FIG:FigPhaseErrorTok} are consistent with our error analysis and simulations. It can be seen that P-BSC~\cite{zhang2024binomial} and I-BSC significantly reduce the ripples induced by motion with the increment of $K$. $K=0$ represents the raw phase without any error compensation. Moreover, I-BSC can better suppress motion error than 3-step and 4-step P-BSC~\cite{zhang2024binomial} when adopting the same binomial order $K$, indicating its fast convergence speed. The nonlinear error compromises the performance of both P-BSC and I-BSC, with the 3-step P-BSC being the most susceptible. Therefore, in the presence of nonlinearity, we recommend performing nonlinearity correction before using BSC, otherwise, choose I-BSC or 4-step P-BSC rather than 3-step P-BSC to have a robust performance. The binomial order $K$ can be flexibly chosen according to the accuracy and time efficiency requirements. In the following real-world experiments, we adopted I-BSC with eight images ($K=4$) to sufficiently suppress both motion error and intensity noise.

\subsection{Computational Efficiency}
\label{SEC:ComputationalEfficiency}
\begin{figure*}[!t]
    \centering
      \begin{minipage}{0.55\linewidth}
            \centering
            \includegraphics[width=1\linewidth]{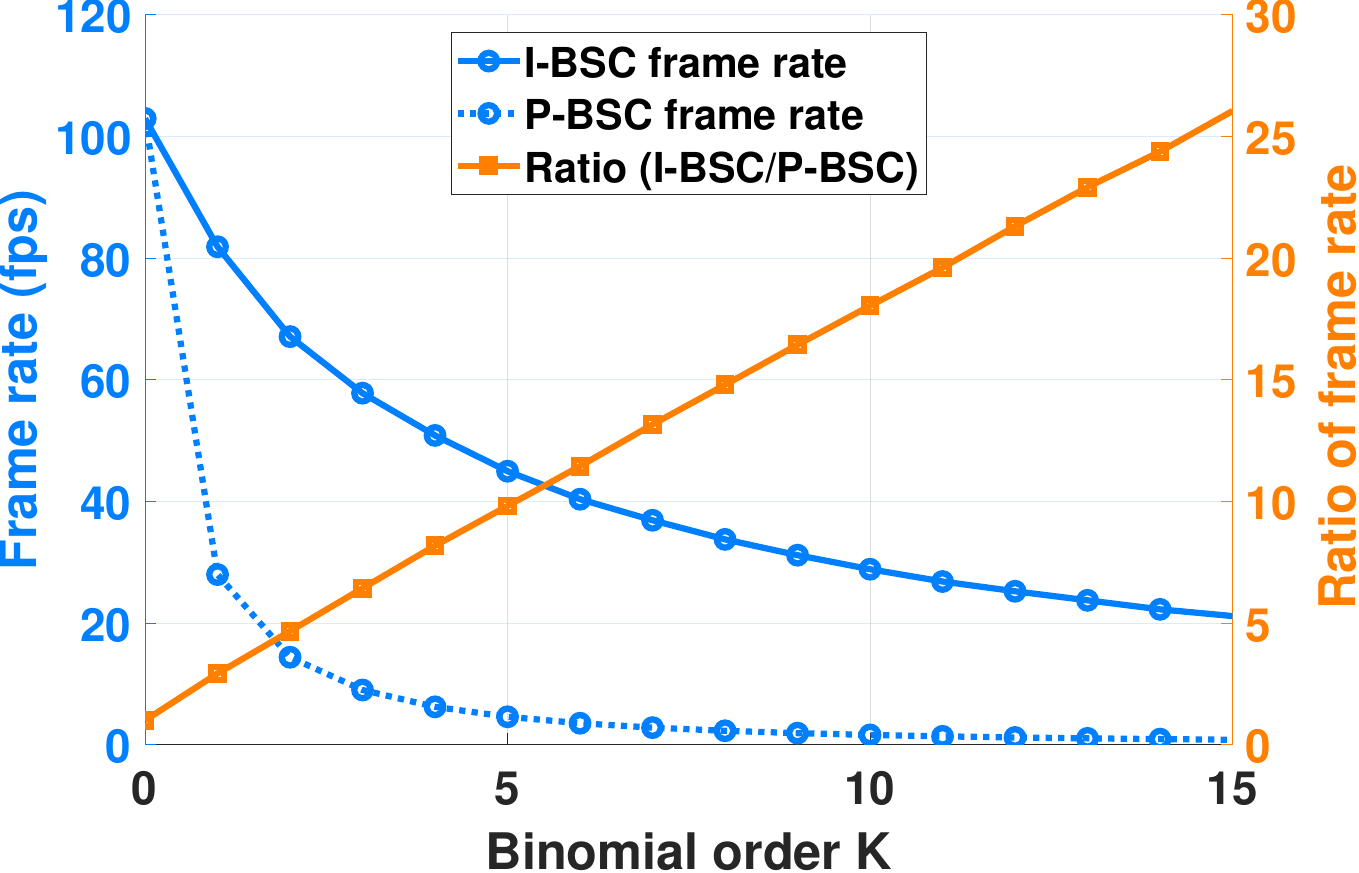}            
      \end{minipage}
     \begin{minipage}{0.4\linewidth}
    \centering
    \renewcommand{\arraystretch}{1.7}
    \begin{tabular}{c|cc}
        \hline
        \textbf{Operation}& \textbf{P-BSC} & \textbf{I-BSC}\\
        \hline
        $\tan^{-1}$ & $K + 1$ & $1$ \\
        \hline
        mod & $(0.5K + 1)(K + 1)$ & $0$ \\
        \hline
        $+/-$ & $(0.5K + 2)(K + 1)$ & $4K + 2$ \\
        \hline
        $\times/\div$ & $K(K + 1)$ & $4K$ \\
        \hline
        cmp$^*$& $0.5K(K + 1)$ & $0$ \\
        \hline
        \multicolumn{3}{r}{\parbox{6cm}{\textit{ All operations are pixel-wise, which means the total number of computation needs to be multiplied by the number of pixels in the image.}}}\\
        \multicolumn{3}{r}{$^*$:\textit{ \parbox{6cm}{Comparison operation.}}} \\
    \end{tabular}
    \end{minipage}
    \caption{Left: Comparison of I-BSC and P-BSC~\cite{zhang2024binomial} in computation frame rate. As the binomial order (number of input images) increases, the speed advantage of I-BSC over P-BSC becomes increasingly significant. Right: Comparing the computation complexity. The codes are written in C++.} 
    \label{FIG:FigSpeed}
\end{figure*}  

\begin{figure}[t!]
    \centering
    \includegraphics[width=0.95\linewidth]{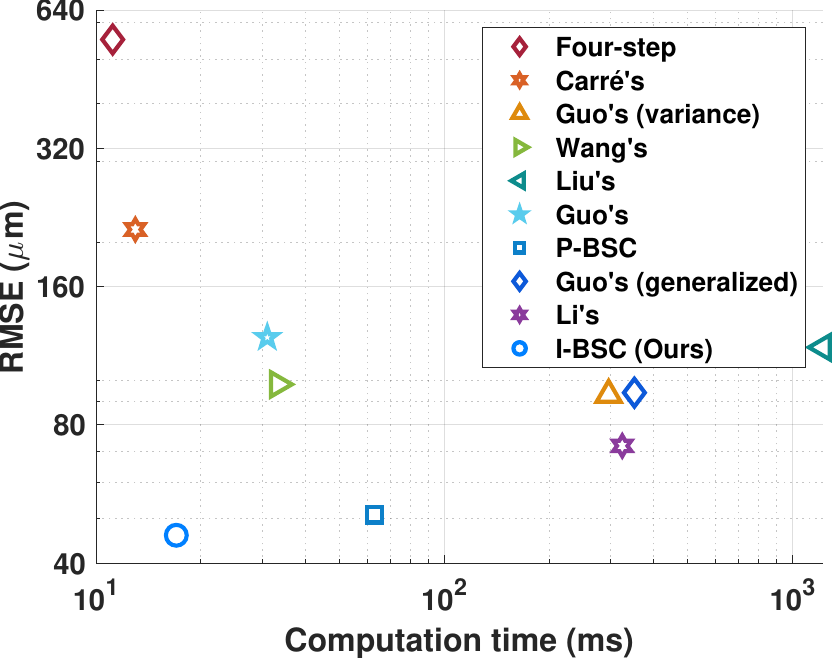}
    \caption{Computation time versus RMSE when measuring a standard plate translation along Z-axis by 1) traditional four-step PSP~(4 frames), 2) Carré's~\cite{carre1966installation}, 3) Guo's~\cite{guo2013phase} (variances)~(4 frmaes), 4) Lu's~\cite{lu2013new}~(4 frames), 5) Wang's~\cite{wang2018motion}~(4 frames), 6) Liu's~\cite{liu2019real}~(8 frames), 7) Guo's~\cite{guo2021real}~(5 frmaes), 8) our previous proposed P-BSC~\cite{zhang2024binomial}~(8 frames), 9) Guo's~\cite{guo2025generalized}~(generalized)~(6 frames), 10) Li's~\cite{li2025mrpca}~(4 frmaes), and Our I-BSC~(8 frames). Our I-BSC achieves the lowest RMSE among all the methods, while it also showcases the shortest computation time except Carré's~\cite{carre1966installation} and traditional four-step methods. The codes are written in Matlab. }
    \label{FIG:FigAccuracyVSTime}
\end{figure}
In terms of computational efficiency, the proposed I-BSC method is characterized by both low complexity and high suitability for parallel acceleration. This advantage originates from its algorithm design, which eliminates the need for iterative procedures or global transformations (e.g., Fourier or Hilbert transforms) over the entire image. Instead, the computation mainly comprises several pixel-wise arithmetic operations—including addition, subtraction, multiplication, and division—along with a single arctangent calculation. As a result, I-BSC introduces only minimal computational overhead compared to the conventional four-step phase-shifting method. 

To benchmark the computational performance, we implemented the proposed I-BSC method (8 frames) and several representative approaches using single-threaded MATLAB. The methods selected for comparison are all designed to compensate for phase deviations. They include: 1) four-step PSP, 2) Carré's~\cite{carre1966installation} (4 frames), 3) Guo's~\cite{guo2013phase} (variance, 4 frames), 4) Wang's~\cite{wang2018motion} (4 frames), 5) Liu's~\cite{liu2019real} (8 frames), 6) Guo's~\cite{guo2021real} (5 frames), 7) P-BSC~\cite{zhang2024binomial} (8 frames), 8) Guo's~\cite{guo2025generalized} (generalized, 6 frames), and 9) Li's~\cite{li2025mrpca} methods (4 frames). It should be noted that only the phase compensation step is considered for all the methods, which takes the image sequence as input and outputs the compensated phase map. We intuitively visualize the performance-complexity trade-offs among the different methods as shown in scatter plot Fig.~\ref{FIG:FigAccuracyVSTime}. We can conclude that our I-BSC demonstrates competitive computational speed compared with traditional four-step methods, yet it attains the lowest RMSE among all the comparison methods.

Further, by avoiding multi-frame phase calculations and layer-by-layer accumulation in P-BSC, our I-BSC exhibits a significant advantage in computational complexity as summarized in Fig.~\ref{FIG:FigSpeed}. Our I-BSC remarkably diminishes all kinds of operations including $\tan^{-1}$, mod, $+/-$, $\times/\div$, and cmp, reducing the time complexity by at least one polynomial order.

To demonstrate the computational speed advantage of I-BSC over P-BSC, we implemented both P-BSC and I-BSC using single-threaded C++ code and ran the programs on an Intel i7-13790F CPU at 2.10 GHz. We conducted experiments for both I-BSC and P-BSC with binomial orders ranging from 0 to 15, performing each operation 100 times. The average computation frame rates are presented in Fig.~\ref{FIG:FigSpeed}. It is evident that I-BSC accelerates P-BSC by several to tens of times, achieving frame rates exceeding 50~fps when $K<5$ even with CPU single-threaded programming. The frame rate improvement increases linearly with the binomial order, indicating that I-BSC reduces the time complexity of P-BSC by one polynomial order. 

Our theoretical analysis, simulations, and experiments show that compared with P-BSC, I-BSC accelerates the computational frame rate by several to a dozen times, while also maintaining faster error convergence. This is attributed to the fact that I-BSC only needs to compute the arctangent function once, avoiding high computational overhead and error accumulation.

\section{Conclusions and Future Outlook}
In conclusion, our analysis separated the confounding motion errors into two distinct components: ghosting artifacts and ripple-like distortions. Based on this, we proposed a robust phase-shifting profilometry for arbitrary motion, including 6-DoF motion, non-rigid deformations, and multi-target movements, achieving high-fidelity motion-error-free 3D reconstruction. We devised a two-stage framework that demonstrated superior motion error suppression performance. It comprised pixel-wise image alignment based on optical flow to address ghosting artifacts, followed by I-BSC to correct ripple-like distortions. We conducted extensive and comprehensive experiments under challenging motion scenarios, including 6-DoF motion (translations and rotations in any direction), non-rigid deformations, and multi-target movements. Experimental results demonstrate that our method delivers robust performance across arbitrary motion scenarios, significantly suppressing motion errors and outperforming SoTA methods.


In the past, PSP was generally considered applicable only to static objects. This notion has gradually been dispelled by emerging researches over the past decade. In this paper, our RPSP-AM extends the applicability of traditional PSP to arbitrary motion scenarios. Examples include dynamic visual localization in robotics, providing micron-level 3D spatial perception for humanoid, industrial, or surgical robots; industrial inspection on the move, enabling continuous 3D inspection on production lines without halting operations; and real-time medical reconstruction, facilitating rapid reconstruction of customized medical devices such as dental models and prosthetics.

In future work, we plan to customize a deep neural network to estimate accurate pixel-wise optical flow directly from fringe images~\cite{lu2023kinematic}, thereby eliminating the need for additional uniform patterns and preserving the frame-wise loopable merit of BSC. In addition, by leveraging the dynamic measurement ability of the proposed RPSP-AM, it can collaborate with 3D Gaussian splatting~\cite{kerbl20233d} for a combined active-passive scanning, thereby achieving high-precision 3D reconstruction of the entire scene from near to far.
\bibliographystyle{ieeetr}
\bibliography{main}
\begin{IEEEbiography}
[{\includegraphics[width=1in,height=1.25in,clip,keepaspectratio]{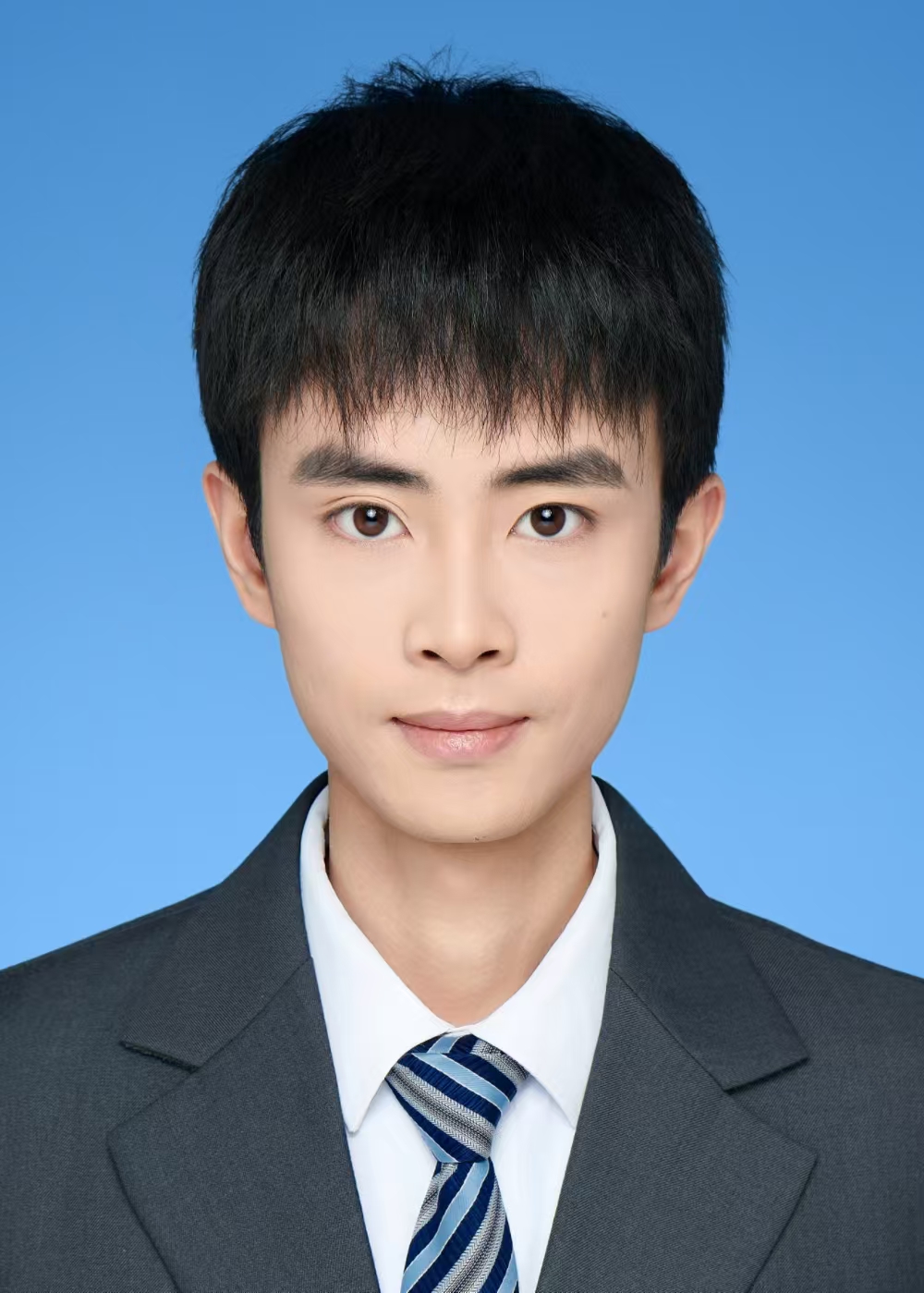}}]{Geyou Zhang} received the B.E. and M.E. degrees from the College of Electrical Engineering at Sichuan University in 2020 and 2023, respectively. He is currently pursuing a Ph.D. in the School of Information and Communication Engineering at the University of Electronic Science and Technology of China. His research interests include 3D imaging, with a focus on structured light illumination, stereo vision, and dynamic 3D scanning. 
\end{IEEEbiography}

\begin{IEEEbiography}
[{\includegraphics[width=1in,height=1.25in,clip,keepaspectratio]{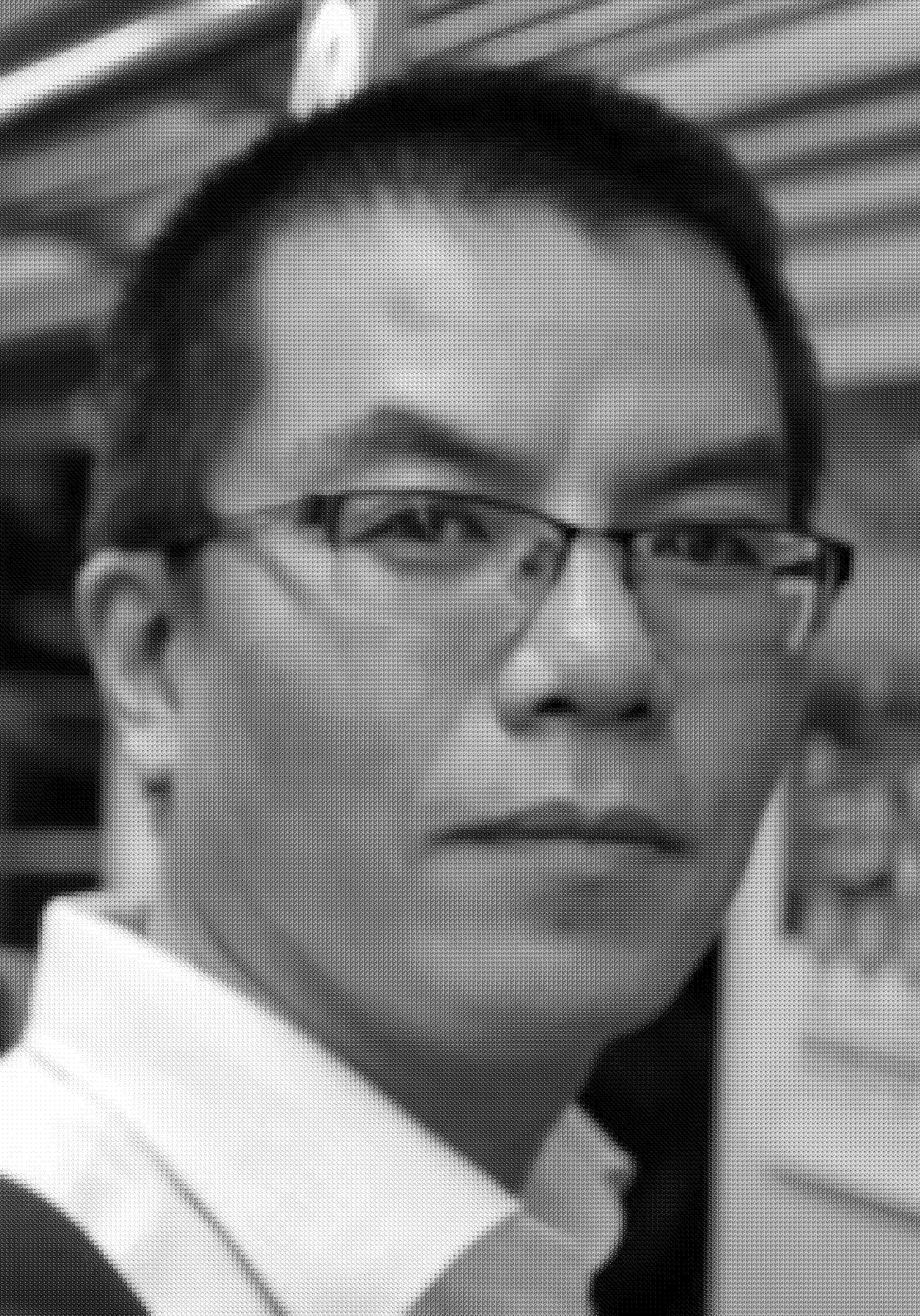}}]{Kai Liu} received the BS and MS degrees in computer science from Sichuan University, China in 1996 and 2001, and the PhD degree in electrical engineering from the University of Kentucky in 2010, respectively. He is currently a professor in the College of Electrical Engineering at Sichuan University. He was a postdoctoral researcher in the Information Access Lab at the University of Delaware from September 2010 to July 2011. His main research interests include computer/machine vision, active/passive stereo vision, and image processing.
\end{IEEEbiography}

\begin{IEEEbiography}
[{\includegraphics[width=1in,height=1.25in,clip,keepaspectratio]{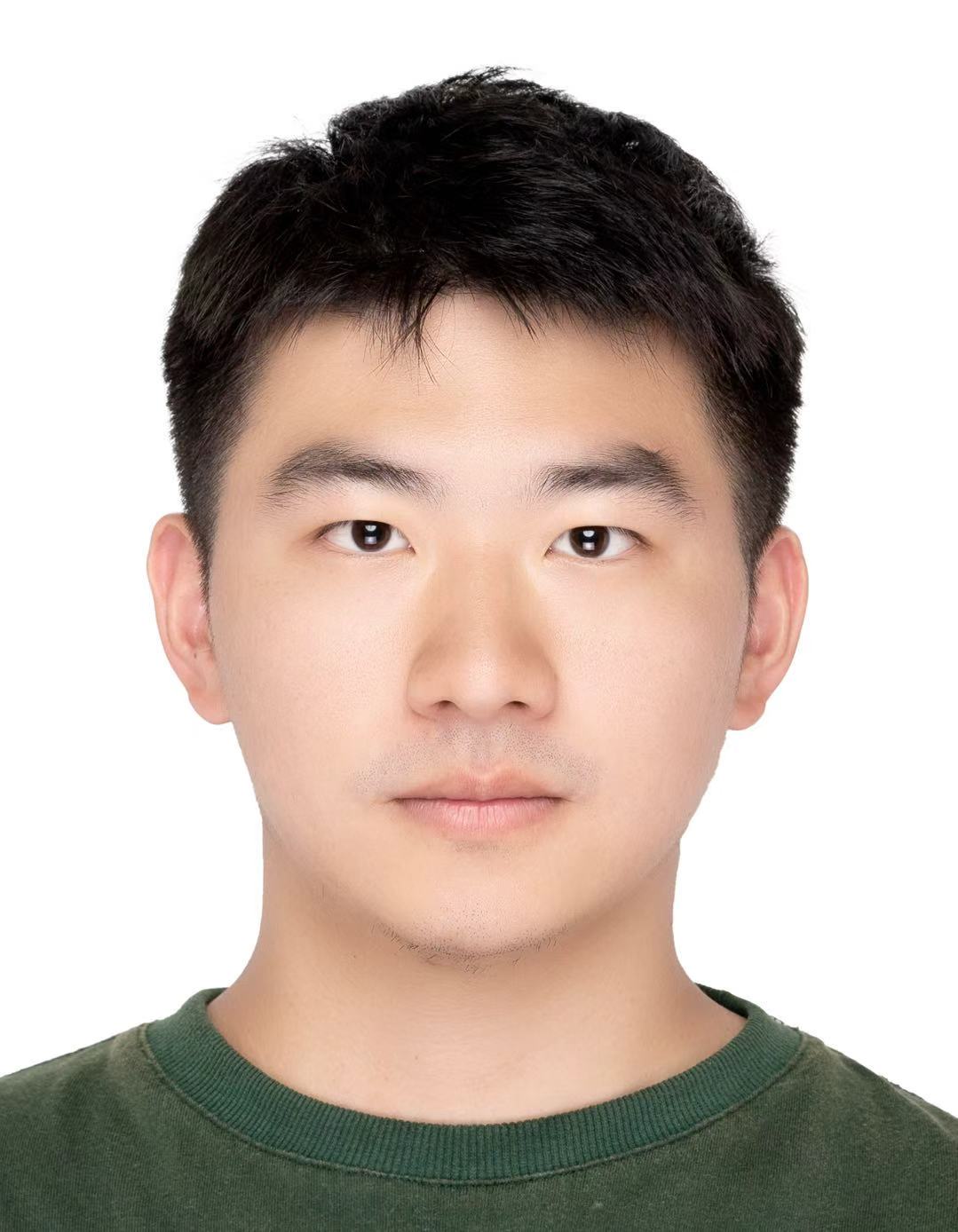}}]{Ao Li} received his B.S. degree from Chongqing University, China, in 2017, and his M.S. degree from Chongqing University of Posts and Telecommunications, China, in 2020. He is currently pursuing a Ph.D. degree in the School of Information and Communication Engineering at the University of Electronic Science and Technology of China. His research interests include deep learning and computer vision.
\end{IEEEbiography}

\begin{IEEEbiography}[{\includegraphics[width=1in,height=1.25in,clip,keepaspectratio]{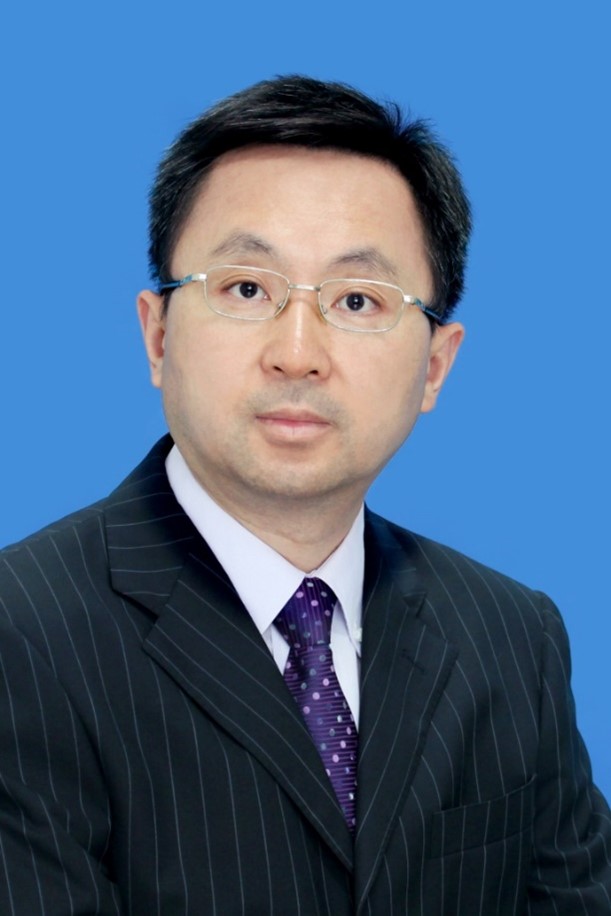}}]{Ce Zhu} (M’03–SM’04–F’17) received the B.S. degree from Sichuan University, Chengdu, China, in 1989, and the M.Eng and Ph.D. degrees from Southeast University, Nanjing, China, in 1992 and 1994, respectively, all in electronic and information engineering. 
He held a post-doctoral research position with the Chinese University of Hong Kong in 1995, the City University of Hong Kong, and the University of Melbourne, Australia, from 1996 to 1998. He was with Nanyang Technological University, Singapore, for 14 years from 1998 to 2012, where he was a Research Fellow, a Program Manager, an Assistant Professor, and then promoted to an Associate Professor in 2005. 
He has been with University of Electronic Science and Technology of China, China, as a Professor since 2012. His research interests include video coding and communications, video analysis and processing, 3D video, visual perception and applications. 

He has served on the editorial boards of a few journals, including as an Associate Editor of IEEE TIP, IEEE TCSVT, IEEE T BROADCAST, IEEE SPL, an Editor of IEEE COMMUN SURV TUT, and an Area Editor of SPIC. He has also served as a Guest Editor of a few special issues in international journals, including as a Guest Editor in the IEEE J-STSP. He is an APSIPA Distinguished Lecturer (2021-2022), and was also an IEEE Distinguished Lecturer of Circuits and Systems Society (2019-2020). He is a co-recipient of multiple paper awards at international conferences, including the most recent Best Demo Award in IEEE MMSP 2022, and the Best Paper Runner-Up Award in IEEE ICME 2020.
\end{IEEEbiography}

\end{document}